\pdfoutput=1
\documentclass[11pt]{article}

\usepackage[final]{acl}
\usepackage{times}
\usepackage{latexsym}

\usepackage[T1]{fontenc}
\usepackage[utf8]{inputenc}

\usepackage{microtype}

\usepackage{inconsolata}

\usepackage{graphicx}

\usepackage{subfigure}
\usepackage{amsmath}
\usepackage{amssymb}
\usepackage{mathtools}
\usepackage{amsthm}
\usepackage{wrapfig}
\usepackage{url}
\usepackage{booktabs}
\usepackage{graphicx}
\usepackage{multirow}
\usepackage{subcaption}
\usepackage{wrapfig}
\usepackage{enumitem}
\title{CheXalign: Preference fine-tuning in chest X-ray interpretation models without human feedback}

\author{
 \textbf{Dennis Hein\textsuperscript{1}},
 \textbf{Zhihong Chen\textsuperscript{1}},
 \textbf{Sophie Ostmeier\textsuperscript{1}},
 \textbf{Justin Xu\textsuperscript{1,2}}, 
 \textbf{Maya Varma\textsuperscript{1}}, \\
 \textbf{Eduardo Pontes Reis\textsuperscript{1}}, 
 \textbf{Arne Edward Michalson\textsuperscript{1}},
 \textbf{Christian Bluethgen\textsuperscript{1,3}}, \\
 \textbf{Hyun Joo Shin\textsuperscript{4}}, 
 \textbf{Curtis Langlotz\textsuperscript{1}},
 \textbf{Akshay S Chaudhari\textsuperscript{1}}
\\
 \textsuperscript{1}Stanford University, USA
 \textsuperscript{2}University of Oxford, UK \\
 \textsuperscript{3}University Hospital Zurich, University of Zurich, Switzerland \\
 \textsuperscript{4}Yongin Severance Hospital, Yonsei University, South Korea
\\
 \small{
     \texttt{heind@stanford.edu}
 }
}

\begin{document}
\maketitle

\begin{abstract}
Radiologists play a crucial role in translating medical images into actionable reports. However, the field faces staffing shortages and increasing workloads. While automated approaches using vision-language models (VLMs) show promise as assistants, they require exceptionally high accuracy. Most current VLMs in radiology rely solely on supervised fine-tuning. Meanwhile, additional preference fine-tuning in the post-training pipeline has become standard practice in the general domain. The challenge in radiology lies in the prohibitive cost of obtaining radiologist feedback at scale. To address this challenge, we propose an automated pipeline for preference feedback, focusing on chest X-ray radiology report generation (RRG). Specifically, our method leverages publicly available datasets containing pairs of images and radiologist-written reference reports with reference-based metrics, or Judges, eliminating the need for \emph{additional radiologist feedback}. We investigate reward overoptimization via length exploitation in this setting and introduce a length-controlled version of the GREEN score. Our best-performing setup achieves state-of-the-art CheXbert scores on the MIMIC-CXR dataset for the RRG task while on average maintaining robust performance across six additional image perception and reasoning tasks.
\end{abstract}

\section{Introduction}
X-rays are one of the most frequently collected imaging studies in clinical practice, with the advantages of wide availability, cost-effectiveness, and low radiation dose. Chest X-rays (CXR) are used for diverse purposes, with approximately 1.4 billion diagnostic X-ray examinations collected per year in the world \citep{paho2012world,world2016communicating,cid2024development}. The amount and significance of CXRs can pose a burden for radiologists and a potential negative impact for patients without timely interpretation, especially for those containing critical lesions \citep{ruutiainen2013increased,hanna2017emergency,bruls2020workload,bhargavan2002too,lyon2015rural,rimmer2017radiologist}.

Recent strides in generative vision-language models (VLMs) hold promising implications for this high-stakes and low-data field \citep{liu2024visual,radford2021learning}. Typically pre-trained using image-text contrastive learning and subsequently fine-tuned, recent VLMs have started to demonstrate promising performance in CXR interpretation and radiology report generation (RRG)~\citep{chen2024chexagent,bannur2024maira2}. In high-stakes fields like radiology, where accurate medical descriptions directly influence disease diagnosis and treatment decisions, the generated outputs must maintain high factual accuracy to ensure patient safety.

However, recent studies have shown that supervised fine-tuning (SFT) might be insufficient in the post-training process. For example,~\citet{hong2024orpo} illustrate the limitation of SFT by training on a preference dataset, containing ``good'' and ``bad'' completions. By tracking the log probabilities of each during the course of training, they show that the log probabilities of the bad completions inadvertently increase alongside the good completions. Preference fine-tuning methods, such as reinforcement learning from human feedback (RLHF)~\citep{ziegler2020finetuning,stiennon2020learning,ouyang2022training}, using Proximal Policy Optimization (PPO)~\citep{schulman2017proximal} or REINFORCE~\citep{williams1992simple}, and direct alignment algorithms (DAAs), such as Direct Preference Optimization (DPO)~\citep{rafailov2023direct}, effectively alleviate this problem by employing a negative gradient to lower probabilities of ``bad'' completions~\citep{tajwar2024preference}. In fact, most recent large language models (LLMs)~\citep{ouyang2022training,bai2022training,touvron2023llama,jiang2024mixtral,anil2024gemini} include some form of preference fine-tuning in their post-training pipeline. Yet, this approach has been only sparsely investigated within the medical vision-language domain (e.g.,~\citet{xiao2024radiology,zhu2025mmedp}), concurrent to our work.

\begin{figure}
    \centering
    \includegraphics[width=\linewidth]{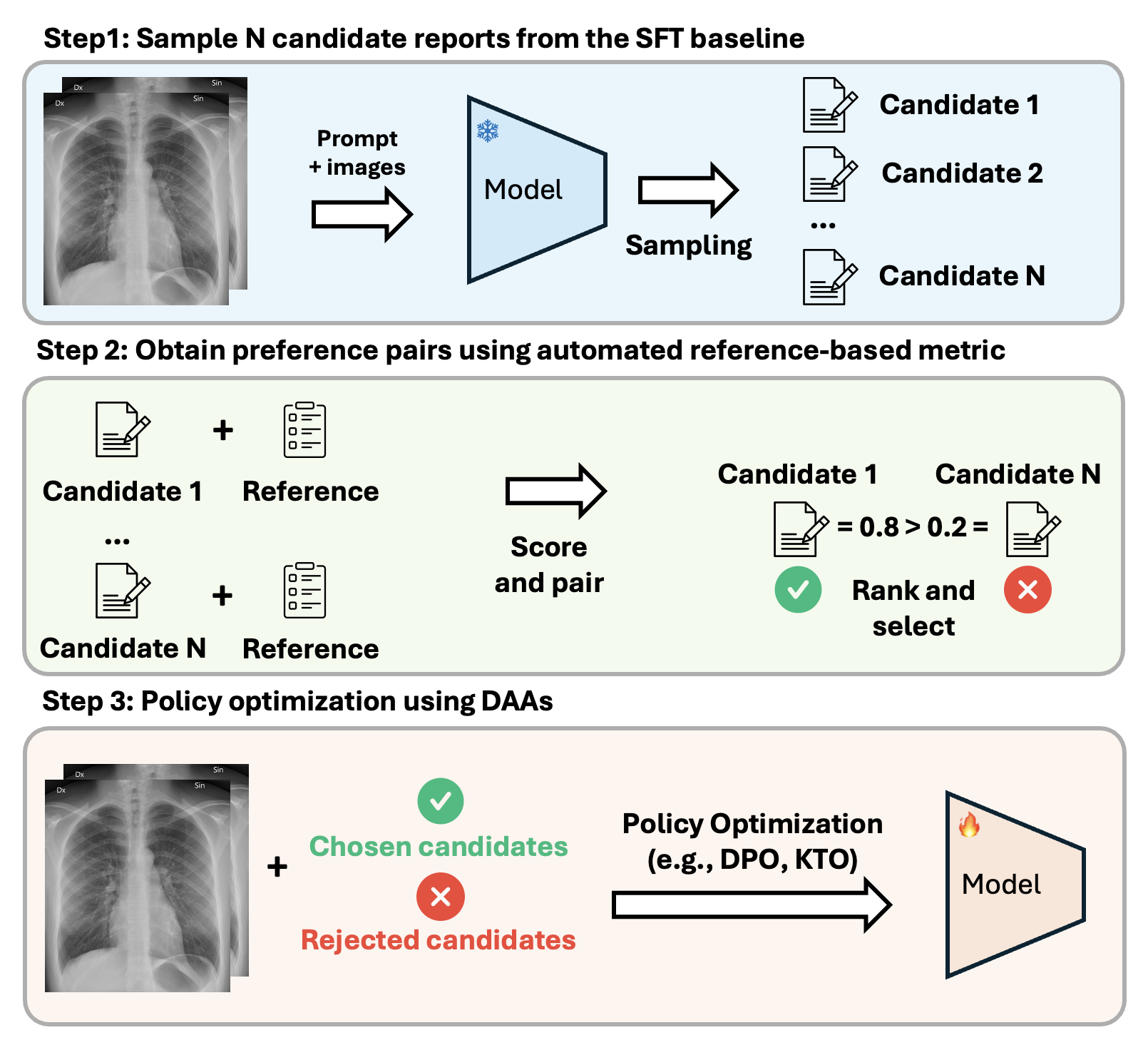}
    \caption{Overview of the CheXalign pipeline. For a given dataset containing CXRs and radiologist written references (e.g., MIMIC-CXR), we obtain preference pairs using some automated reference-based metric (e.g., GREEN or the BERTScore), and then optimize the policy using light-weight DAAs (e.g., DPO or KTO).}
    \label{fig:overview}
\end{figure}

The primary challenge hindering the application of preference fine-tuning in the post-training of VLMs in radiology is the \emph{prohibitive cost} of obtaining radiologist preferences at scale. To overcome this obstacle, we introduce CheXalign, an automated pipeline for generating preference data for the crucial RRG task. 
Specifically, we leverage the availability of reference reports written by radiologists in a clinical setting within large, publicly available, datasets such as MIMIC-CXR~\citep{johnson2019mimic} and CheXpert Plus~\citep{chambon2024chexpertplus}. This allows us to use reference-based, uni-modal,  metrics, such as GREEN~\citep{ostmeier2024green}, a recent state-of-the-art LLM-based metric for evaluating CXR reports, to annotate generated reports in a factually grounded fashion. An overview of our preference fine-tuning pipeline is available in Fig. \ref{fig:overview}. Our approach enables us to obtain high-quality preference datasets in a fully automated and scalable manner. Using our proposed method, we systematically study how DAAs can be used to enhance the clinical efficacy of medical VLMs \emph{without any additional radiologist feedback}. Our contributions are as follows:
\begin{enumerate}
\item We introduce an automated pipeline for preference pair generation in RRG models, circumventing the \emph{prohibitively expensive} task of obtaining preference feedback from radiologists at scale. 
\item We systematically evaluate and benchmark the proposed pipeline using different reference-based metrics, DAAs, and RRG models. Our findings indicate that the RRG performance can be improved even when using inexpensive, general domain, natural language generation (NLG) metrics for preference pair generation. 
\item Using our proposed pipeline, we obtain new state-of-the-art CheXbert scores on the MIMIC-CXR data for the RRG task. 
\item We study reward overoptimization via length exploitation, and introduce the length-controlled GREEN score. 
\item We benchmark our models post alignment on set of diverse additional image perception and reasoning tasks to assess whether there is an alignment tax in this setting.
\end{enumerate}
Code for this project is available in the following repository: \url{https://github.com/StanfordMIMI/CheXalign}.

\section{Related Works}
VLMs~\citep{radford2021learning,li2021albef,li2022blip,li2023blip2,liu2024visual} are a multi-modal extension to LLMs. In this setting, the prompt $x$ contains images and/or text. Typical tasks include Vision Question Answering (VQA) and image captioning (e.g., RRG in the field of radiology). There is also a line of works to extend VLMs to the medical domain~\citep{thawkar2023xraygpt,hyland2023maira,chaves2024llavarad,tu2024towards,bannur2024maira2,chen2024chexagent,chen2024chexagent2,lee2024llmcxr,jin2024prompmrg} which mainly focus on CXR interpretation and RRG due to the wide availability of public datasets~\citep{johnson2019mimic,chambon2024chexpertplus}. However, even with strong LLMs and vision-backbones, VLMs have been observed to ``hallucinate'' and produce outputs that are not factually grounded in the image~\citep{zhou2024aligning}. Such hallucinations represent a significant risk in high-stakes healthcare fields such as radiology. Similar to~\citet{zhou2024aligning}, we pose the problem of hallucinations as an alignment problem and propose tackling it via preference fine-tuning. As noted above, preference fine-tuning, although now a standard part of the post-training pipeline for general domain LLMs, remains largely unexplored in the context of medical VLMs.

RLHF~\citep{ziegler2020finetuning,stiennon2020learning,ouyang2022training} is a very effective technique for aligning LLMs and VLMs with human preferences. It was instrumental in the development of frontier models such as ChatGPT and Bard~\citep{lee2024rlaif}. RLHF involves two stages: reward modeling, where a reward model is learned on the preference data, and alignment, where reinforcement learning (RL) algorithms are used to optimized the proxy reward. While extremely effective, RLHF is very computationally heavy and can be finicky for non-experts. Relatively recently, a new class of algorithms called DAAs~\citep{rafailov2024scaling} have become increasingly popular.\footnote{Used more loosely than in ~\citet{rafailov2024scaling}.} This class of algorithms re-parameterize the reward model via a change-of-variables using the closed-formed solution to the RLHF objective, effectively bypassing both the reward modeling and RL stages. This yields alignment algorithms that remain performant yet computationally more lightweight and significantly easier to implement. DPO~\citep{rafailov2023direct} was the first in this category and remains one of the most popular versions. After the advent of DPO, a large  variety of DAAs have been suggested \citep{azar2023general, park2024disentangling, ethayarajh2024kto, hong2024orpo}. A brief introduction to RLHF and DAAs is available in \S\ref{appenidx:daas}. Another dimension for improved efficiency is reinforcement learning from AI feedback (RLAIF), first introduced in~\citet{bai2022constitutional}. Replacing humans with LLMs leads to significant reductions in cost, making it much more scalable, while maintaining high quality~\cite{lee2024rlaif}. In this work, we will leverage these advancements for scalable, fully automated, preference data generation and computationally lightweight alignment. 

Since the reward model in the RLHF objective is learned, it is an imperfect proxy of the ground truth reward. As this proxy is optimized, ground truth performance might saturate or even deteriorate.\footnote{As per Goodhart's law: “When a measure becomes a target, it ceases to be a good measure.”~\citep{gao2023scaling}.} This reward overoptimization, or hacking, phenomena was first studied in~\citet{gao2023scaling} for RLHF. Despite not fitting an explicit reward model, similar behavior has been observed empirically for DAAs~\citep{rafailov2024scaling}. In particular, length exploitation, the tendency to learn to produce excessively verbose completions, is one common dimension of reward overoptimization, observed in both RLHF and for DAAs. For instance,~\citet{park2024disentangling} showed that DPO amplifies minor verbosity bias embedded in the preference data. In this work, we explore this phenomenon in the context of preference fine-tuning of RRG models.

\section{Methodology} 
\subsection{RRG Preference Fine-tuning without Human Feedback}
Expert human feedback from radiologists is the gold standard for preference data generation and evaluation for the RRG task. However, scaling is impractical, if not unfeasible, due to the limited availability of radiologists for large-scale annotation tasks. In the general domain, it is common to leverage LLMs for cost effective preference data generation~\citep{bai2022constitutional,dubois2023alpacafarm,lee2024rlaif}.~\citet{zheng2023judging} categorized ``LLM-as-a-Judge'' evaluation methods into pairwise, single answer, and reference-guided grading. Pairwise grading being the most common in the general domain both for preference data generation~\citep{dubois2023alpacafarm,lee2024rlaif} and evaluation~\citep{zheng2023judging,dubois2024lengthcontrolled}. 

\begin{figure*}
    \centering
    \includegraphics[width=\linewidth]{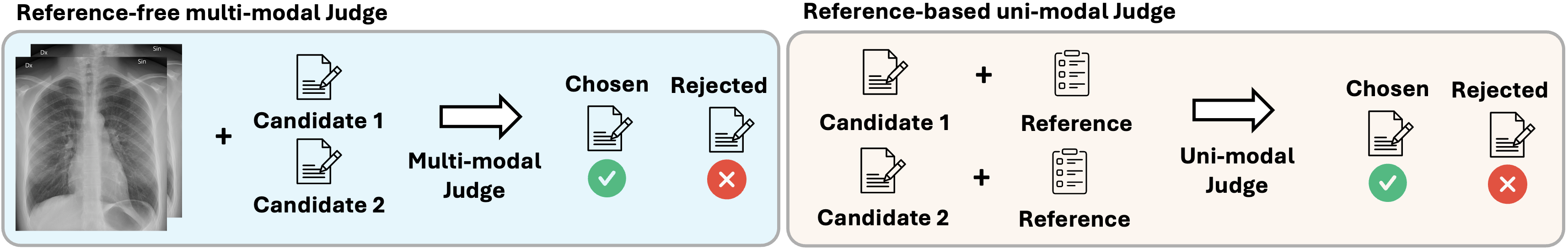}
    \caption{Illustration of preference pairs generation using a reference-free multi-modal Judge (left) and a reference-based uni-modal Judge (right). Human (radiologist) feedback is multi-modal in nature but prohibitively expensive at scale. On the other hand, using LLM, or metrics-based, Judges is highly scalable. However, a high-quality multi-modal Judge is difficult to obtain. For the RRG task, we propose leveraging large publicly available datasets, containing CXRs and radiologist written reference reports, as this enables scalable, factually grounded, preference data generation using reference-based metrics.}
    \label{fig:reference_compare}
\end{figure*}

These existing methods, however, are tailored for uni-modal, general-domain LLMs and do not directly apply to our multi-modal setting, which involves both visual and textual data. Moreover, factual grounding is essential in RRG to ensure clinical reliability. To overcome these challenges, we propose using reference-based grading, leveraging publicly available datasets that contain paired prompts---including images---and \emph{radiologist-written} reference reports. This abundance of high-quality references allows us to provide factually grounded annotations without the need for a multi-modal metrics, or ``Judge'', setting our approach apart from prior studies of preference alignment of VLMs, such as~\citet{sun2024aligning}. A comparison of reference-free and reference-based metrics, or ``Judges'', in this setting is available in Fig.~\ref{fig:reference_compare}. For a given Judge, we obtain preference pairs by repeat sampling from the SFT baseline. Canonical alignment algorithms, such as DPO, can then be used to preference fine-tune the model. 

\subsection{Evaluation}
Since we have radiologist-written reference available, it is possible to employ standard, general domain, NLG metrics such as BLEU~\citep{papineni2002bleu}, ROUGE~\citep{lin2004rouge}, and the BERTScore~\citep{zhang2019bertscore}. However, these NLG metrics may not be able to differentiate between subtle nuances that are clinically relevant. Thus, we include two validated, radiology-specific, and clinically relevant, metrics: GREEN~\citep{ostmeier2024green} and CheXbert scores~\citep{smit2020combining}. GREEN is a state-of-the-art metric for radiology report evaluation, based on a single-answer reference-guided LLM-as-a-Judge mechanism. CheXbert scores are a clinical efficacy metrics based on extracting 14 labels\footnote{Enlarged Cardiomediastinum, Cardiomegaly, Lung Opacity, Lung Lesion, Edema, Consolidation, Pneumonia, Atelectasis, Pleural Effusion, Pneumothorax, Pleural Other, Fracture, Support Devices, No Finding.} using the CheXbert labeler~\citep{smit2020combining} from the candidate and reference reports.

Initial experimentation illustrated that reward hacking via length exploitation might be a concern in our setting when using the GREEN score as Judge. To counteract this, we propose a simple heuristic approach to explicitly control for length:
\begin{equation*}
    \text{LC-GREEN} := \text{GREEN} / \max (\text{rel\_verbosity}, 1),
\end{equation*}
where $\text{rel\_verbosity}$ denotes the relative verbosity (length in words) of the candidate report compared to the reference report. We call this metric length-controlled GREEN (LC-GREEN).

\section{Experimental Details} 
\subsection{SFT Baselines}
We adopt CheXagent~\citep{chen2024chexagent} (8B) as a representative example of a state-of-the-art, open source, English language, VLM for the RRG task. It has been trained in a canonical way by first adapting an LLM, Mistral-7B~\citep{jiang2023mistral}, to medical text by continued pre-training. Second, a vision encoder, EVA-CLIP-g~\citep{sun2023evaclip}, was adapted via vision pre-training, using contrastive learning on CXR image-text pairs. Third, the two modalities were merged by training a vision-language bridger, or adapter network, keeping the LLM and vision encoder frozen. Finally, the model was instruction-tuned for a range of tasks, including RRG. 
Concurrent to this work, CheXagent has been greatly improved upon with CheXagent-2~\citep{chen2024chexagent2} (3B). CheXagent-2 adopted a fine-tuned SigLIP~\cite{zhai2023siglip} model\footnote{\url{https://huggingface.co/StanfordAIMI/XraySigLIP__vit-l-16-siglip-384__webli}.} as the vision encoder and a fine-tuned Phi-2~\cite{li2023phi2} model\footnote{\url{https://huggingface.co/StanfordAIMI/RadPhi-2}.} (2.7B) as the language decoder. For the image-text connector, instead of using the attention layer as in CheXagent, it uses LLaVA-style MLP connector~\cite{liu2024llava}. CheXagent-2 additionally use indications, which provide clinical context to the patient being imaged, to aid in the RRG task. 

\subsection{Datasets}
\label{dataset}
We use the MIMIC-CXR~\citep{johnson2019mimic} dataset for training, validation and testing. The image-report pairs consist of one or two CXRs and the corresponding free-text findings section. CheXagent-2 additionally uses the indications. For CheXagent, we randomly sample 80k examples as our training data. For CheXagent-2, we opted to use the full MIMIC-CXR training set\footnote{As can be seen below, CheXagent-2 is a strong SFT baseline and we hypothesized that a larger preference dataset would be required to obtain significant performance gains.} (148k examples). To test robustness for the RRG task, we additionally include test data from the CheXpert Plus~\citep{chambon2024chexpertplus} dataset. Moreover, to evaluate whether there is an alignment tax, we additionally evaluate our aligned models on six additional CXR tasks: view classification, binary image classification, single disease identification, multi disease identification, VQA, and image-text reasoning, using test data from five additional datasets RSNA~\citep{shih2019rsna}, SIIM~\citep{siim}, OpenI~\citep{demner2016openi}, SLAKE~\citep{liu2021slake}, and Rad-Restruct~\citep{pellegrini2023radrestruct} datasets. All datasets are in English. 

\subsection{Preference Data}
\begin{table*}[h]
    \centering
    \resizebox{0.9\linewidth}{!}{%
    \begin{tabular}{@{}llcccc@{}}
\toprule
\multirow{2}{*}{\textbf{Algorithm}} & \multicolumn{1}{c}{\multirow{2}{*}{\textbf{Objective}}}                                                                                                                                                                            & \textbf{Preference}           & \multirow{2}{*}{\textbf{Reference}} & \textbf{Length}           & \textbf{Relative}        \\
                                    & \multicolumn{1}{c}{}                                                                                                                                                                                                               & \textbf{pairs}                &                                     & \textbf{controlled}       & \textbf{wall-clock time} \\ \midrule
DPO                                 & $-\log \sigma\left(\beta \log \frac{\pi_\theta\left(y_c \mid x\right)}{\pi_{\mathrm{ref}}\left(y_c \mid x\right)}-\beta \log \frac{\pi_\theta\left(y_r \mid x\right)}{\pi_{\mathrm{ref}}\left(y_r \mid x\right)}\right)$           & $\checkmark$                  & $\checkmark$                        & $\times$                  & 1.0                      \\
LC-DPO                                 & $-\log \sigma\left(\beta \log \frac{\pi_\theta\left(y_c \mid x\right)}{\pi_{\mathrm{ref}}\left(y_c \mid x\right)}-\beta \log \frac{\pi_\theta\left(y_r \mid x\right)}{\pi_{\mathrm{ref}}\left(y_r \mid x\right)}+\alpha(|y_c|-|y_r|)\right), \alpha>0$           & $\checkmark$                  & $\checkmark$                        & $\checkmark$                  & 1.0                      \\
IPO                                 & $ \left( \log \frac{\pi_\theta(y_c|x)}{\pi_{\text{ref}}(y_c|x)} - \log \frac{\pi_\theta(y_r|x)}{\pi_{\text{ref}}(y_r|x)} - \frac{1}{2\tau} \right)^2 $                                                                             & $\checkmark$                  & $\checkmark$                        & $\times$                  & 1.0                      \\
\multirow{2}{*}{KTO}                & $-\lambda_c \sigma \left( \beta \log \frac{\pi_\theta(y_c|x)}{\pi_{\text{ref}}(y_c|x)} - z_{\text{ref}} \right) +  \lambda_r \sigma \left( z_{\text{ref}} - \beta \log \frac{\pi_\theta(y_r|x)}{\pi_{\text{ref}}(y_r|x)} \right),$ & \multirow{2}{*}{$\times$}     & \multirow{2}{*}{$\checkmark$}       & \multirow{2}{*}{$\times$} & \multirow{2}{*}{2.2}     \\
                                    & $\text{where} \,\, z_{\text{ref}}=\mathbb{E}_{(x,y)\sim \mathcal{D}}[\beta  \mathbb{D}_{\text{KL}} \left(\pi_{\theta}(y|x)||\pi_{\text{ref}}(y|x)\right)]$                                                                         &                               &                                     &                           &                          \\
\multirow{2}{*}{ORPO}               & $-\log p_\theta(y_c|x) - \lambda  \log \sigma \left(\log \frac{p_\theta(y_c|x)}{1 - p_\theta(y_c|x)} - \log \frac{p_\theta(y_r|x)}{1 - p_\theta(y_r|x)}  \right),$                                                                 & \multirow{2}{*}{$\checkmark$} & \multirow{2}{*}{$\times$}           & \multirow{2}{*}{$\times$} & \multirow{2}{*}{0.7}     \\
                                    & $\text{where} \,\, p_\theta(y|x) = \exp\left( \frac{1}{|y|} \log \pi_\theta(y|x) \right)$                                                                                                                                          &                               &                                     &                           &                          \\ \bottomrule
\end{tabular}
    }
    \caption{Overview of the DAAs considered in this paper. Preference pairs indicates whether the method requires paired data of chosen/rejected or only binary feedback indicating whether a completion is desirable/undesirable. Reference indicates whether an additional reference model is loaded during training. Length controlled indicates whether the objective directly controls for the length of the completions in order to mitigate reward hacking via length exploitation. Relative wall-clock time is measured as the total time to train one epoch, relative to DPO.}
    \label{tab:preference_optimization}
\end{table*}

\label{preferencedata}
We evaluate two reference-based Judges: GREEN~\citep{ostmeier2024green} and the BERTScore~\citep{zhang2019bertscore}. We obtain our preference data as follows: 1) for each example in the training data, we prompt the SFT baseline $N=4$ times; 2) we get the score for each of the generated reports, compared with the corresponding singular reference; 3) we set the chosen and rejected completions as the highest and lowest scores, omitting the observation if all $N=4$ scores are equivalent. For CheXagent, this rejection rule results in the rejection of 1,246 (1.6\%)~examples for GREEN and 31 (0.04\%)~examples for the BERTScore. This is similar for CheXagent-2, with 3744~(2.53\%)~examples and 147~(0.10\%)~examples rejected for GREEN and the BERTScore, respectively. Summary statistics for the chosen and rejected subsets are available in Table \ref{tab:reward_summary}. We also report summary statistics of the length (in words) of the generated reports. Notably, the spread in average length of the chosen and rejected subsets is slightly more pronounced for GREEN than for the BERTScore, 6.9\% compared to 5.8\% and much more significant overall for CheXagent-2 with a difference of 19.3\% and 17.1\%, respectively. Examples from the chosen and rejected subsets are available in Fig.~\ref{fig:chosen_rejected_chexagent} and~\ref{fig:chosen_rejected_chexagent2}.

\label{results}
\begin{figure*}[!htbp]
    \centering
    \begin{subfigure}
        \centering
        \includegraphics[width=0.49\textwidth]{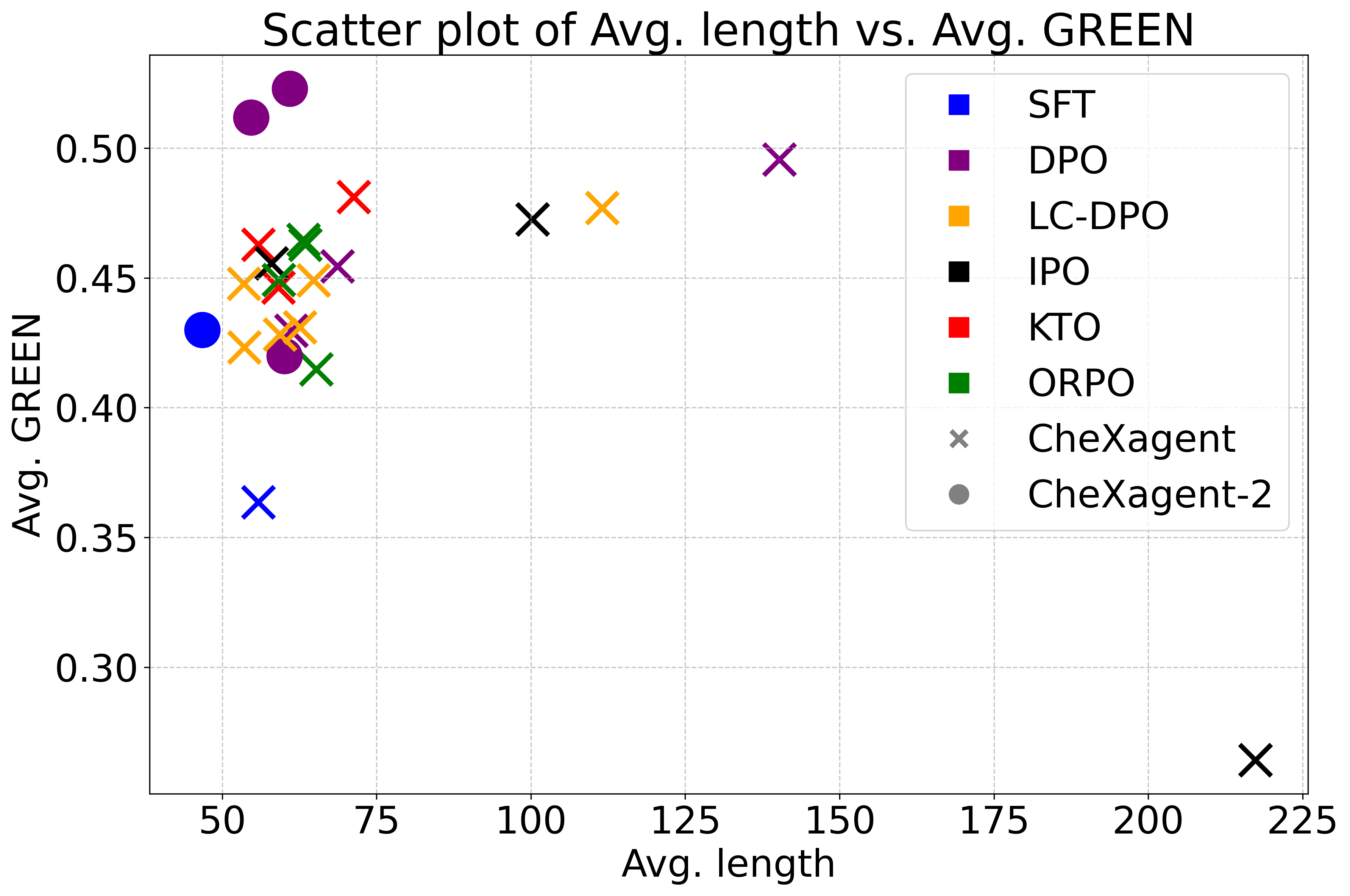}
    \end{subfigure}
    \hfill
    \begin{subfigure}
        \centering
        \includegraphics[width=0.49\textwidth]{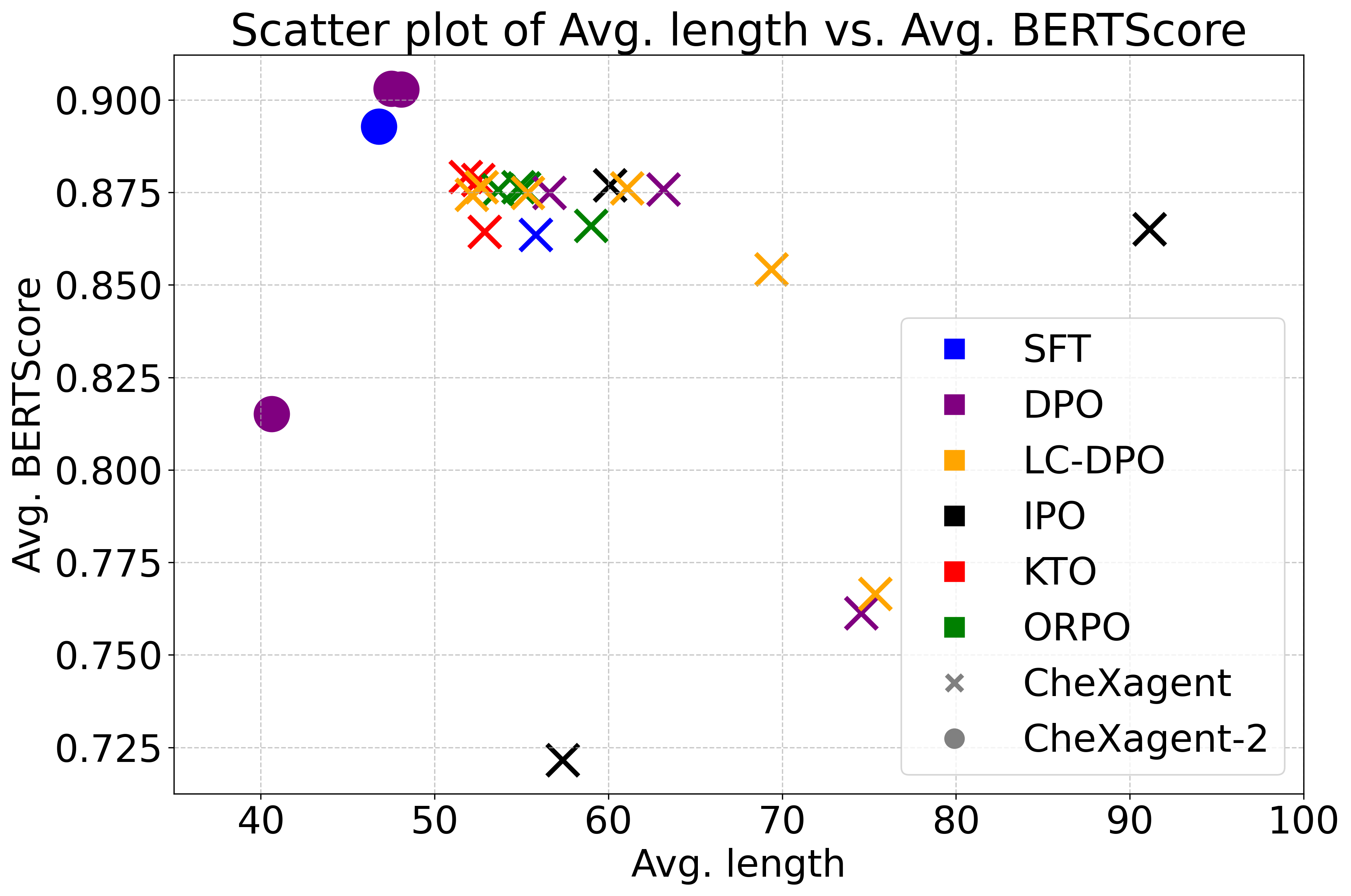}
    \end{subfigure}
    \caption{Average length against average GREEN and BERTScore for all aligned policies on the MIMIC-CXR validation set using GREEN (left) and the BERTScore (right) as Judges.}
    \label{fig:reward_length_scatter}
\end{figure*}

\subsection{Alignment Algorithms}
We opt for a representative subset of available, offline, DAAs. DPO is the original DAA and serves as our baseline. In addition to DPO, we consider: 1) Length-controlled DPO (LC-DPO)~\citep{park2024disentangling}, as an example of a DAA with explicit length regularization. LC-DPO is an elegant extension of DPO, with an additional hyperparameter $\alpha$ which controls the strength of an additional length regularization term. Setting $\alpha=0$ reverts the objective to that in DPO. 2) Identity Preference Optimization (IPO)~\citep{azar2023general} as an example of a DAA with generalized preference, relaxing the assumption of the Bradley-Terry model used in DPO. The authors argue that this helps mitigate over-fitting issues even when preferences are transitive. Relatively recent work has shown that IPO indeed seems to be less prone to reward overoptimization~\citep{rafailov2024scaling}. 3) Kahneman-Tversky optimization (KTO)~\citep{ethayarajh2024kto} as an example of a DAA that does not require preference pairs, but instead only binary feedback on whether a completion is desirable or undesirable. This type of data is much more ubiquitous in practice. In addition, for any given dataset of preference pairs, KTO provides twice the number of examples. 4) Odds-Ratio Preference Optimization (ORPO)~\citep{hong2024orpo}, almost outside of the definition of DAAs, is not based on the RLHF objective but instead appends an additional penalty directly to the negative log likelihood used in SFT. This adds a ``negative gradient'', using the terminology in~\citet{tajwar2024preference}, which will help reduce the log probabilities of rejected completions. 

An overview of all DAAs considered in this paper is available in Table~\ref{tab:preference_optimization}. More implementation details are available in \S\ref{appendix:implementation}.\footnote{Due to compute constraints, we only include results for the baseline alignment algorithm, DPO, for CheXagent-2.}

\section{Results and Analyses}
\subsection{Length Exploitation}
We investigate the issue of length exploitation by plotting the average GREEN and BERTScore against average lengths of the resulting radiology reports using GREEN and the BERTScore as Judges for preference data generation. Results for single runs on the MIMIC-CXR validation set, including all configurations considered, are available in Fig.~\ref{fig:reward_length_scatter}. Save for a IPO run in the lower right corner, there seems to be a positive correlation between average GREEN and average length for CheXagent. In fact, for all DAAs, except for ORPO, there is an indication for a trade-off between added verbosity and GREEN. In addition, such a trade-off is not observed for the BERTScore, shown to the right in Fig.~\ref{fig:reward_length_scatter}. Except for a very tight clustering around 0.87-0.88, there are no clear trends.

\begin{table}[t]
%\begin{wraptable}{r}{0.5\textwidth}
\centering
\resizebox{\linewidth}{!}{%
\begin{tabular}{l*{4}{c}}
\toprule
& \multicolumn{2}{c}{\textbf{GREEN}} & \multicolumn{2}{c}{\textbf{LC-GREEN}} \\
\cmidrule(lr){2-3} \cmidrule(lr){4-5} 
\textbf{Method} & \textbf{Avg. length} & \textbf{Rel. verbosity} & \textbf{Avg. length}  & \textbf{Rel. verbosity} \\ \midrule 

CheXagent	&55.8	&	&55.8	& \\
+DPO	&140.2	&2.51	&68.7	&1.23 \\
+LC-DPO	&111.5	&2.00	&53.5	&0.96 \\
+IPO	&100.2	&1.80	&58.0	&1.04 \\
+KTO	&71.3	&1.28	&55.9	&1.00 \\
+ORPO	&63.1	&1.13	&63.1	&1.13 \\ \midrule 
CheXagent-2	&46.8	&	&46.8	& \\ 
+DPO	&60.9	&1.30	&54.7	&1.17 \\ \midrule 
Reference	&58.4	&1.05	&58.4	&1.05 \\ 
\bottomrule
\end{tabular}
}
\caption{Average length of reports on the MIMIC-CXR validation set of best performing configurations according to GREEN and GREEN-LC, respectively. Relative verbosity is relative to the SFT baseline.}
\label{tab:avg_length}
%\end{wraptable}
\end{table}

\begin{table*}[!htbp] %!htbp
\centering
\resizebox{0.8\linewidth}{!}{%
\begin{tabular}{l*{7}{c}}
\toprule
& & \multicolumn{3}{c}{\textbf{MIMIC-CXR}} & \multicolumn{3}{c}{\textbf{CheXpert Plus}} \\
\cmidrule(lr){3-5}  \cmidrule(lr){6-8}  
\textbf{Model} & \textbf{Judge} & \textbf{GREEN}~$(\uparrow)$  & \textbf{LC-GREEN}~$(\uparrow)$ & \textbf{BERTScore}$(\uparrow)$ & \textbf{GREEN}~$(\uparrow)$  & \textbf{LC-GREEN}~$(\uparrow)$ & \textbf{BERTScore}$(\uparrow)$  \\ \midrule 
CheXagent	&	&0.249		&0.218	&	0.856	& 0.248	&	0.202	 &0.851	\\ \midrule 
+DPO	& & 0.325	(30.6)	& 0.263	(20.5)	&0.862	(0.65)	&0.322	(29.9)	&0.222	(9.69)	&0.859	(0.91)	\\
+LC-DPO	& & 0.320	(28.7)	& 0.288	(32.0)	&0.864	(0.90)	 &0.330	(32.8)	&\textbf{0.268	(32.3)}	&0.861	(1.26)	 \\
+IPO	& GREEN & 0.326	(30.9)	& 0.282	(29.3)	&0.863	(0.84) &0.334	(34.6)	&0.267	(31.9)	&0.861	(1.15)	 \\
+KTO	& & \textbf{0.328	(31.9)}	& \textbf{0.293	(34.1)}	&0.867	(1.27)	 &\textbf{0.341	(37.2)}	&0.266	(31.4)	&0.863	(1.42)	 \\
+ORPO	& & 0.322	(29.4)	& 0.275	(26.2)	&0.862	(0.69)	 &0.326	(31.4)	&0.232	(14.6)	&0.856	(0.64)	\\  \midrule % 
+DPO	& & 0.285	(14.4)	& 0.242	(11.0)	&0.869	(1.56)	  &0.278	(12.0)	&0.208	(2.84)	&0.867	(1.89)	\\
+LC-DPO	& & 0.283	(13.5)	& 0.259	(18.7)	&0.871	(1.75)	 &0.299	(20.3)	&0.248	(22.5)	&\textbf{0.871	(2.34)} \\
+IPO	& BERTScore & 0.283	(13.6)	& 0.247	(13.1)	&0.870	(1.62) &0.282	(13.6)	&0.218	(7.88)	&0.866	(1.75) \\
+KTO	& & 0.304	(21.9)	& 0.279	(28.2)	&\textbf{0.872	(1.88)}	 &0.308	(24.0)	&0.249	(23.1)	&0.869	(2.11)	\\
+ORPO	& & 0.291	(16.9)	& 0.265	(21.4)	&0.869	(1.57)   &0.298	(19.9)	&0.240	(18.4)	&0.870	(2.21)	\\ \midrule 
CheXagent-2	&	&0.326  &0.297		&0.888 &0.349		 &0.304	&0.892\\ \midrule 
+DPO	& GREEN & 	\textbf{0.387	(18.9)}	&\textbf{0.339	(14.1)}	&0.891	(0.30) & \textbf{0.387	(10.9)}	&\textbf{0.320	(5.34)}	&0.888	(-0.38)\\ 
+DPO	& BERTScore & 	0.352	(8.11)	&0.326	(9.58)	&\textbf{0.896	(0.95)}  & 0.359	(2.88)	&0.310	(1.78)	&\textbf{0.893	(0.15)}\\
\bottomrule
\end{tabular}
}
\caption{Results on the MIMIC-CXR and CheXpert Plus test sets (percentage change compared to SFT baseline in brackets). Best results in bold, separate for CheXagent and CheXagent-2.}
\label{tab:nxm}
\end{table*}

\begin{table*}[h]
\centering
\resizebox{0.65\linewidth}{!}{%
\begin{tabular}{l*{6}{c}}
\toprule
&& \multicolumn{2}{c}{\textbf{F1-14}} & \multicolumn{2}{c}{\textbf{F1-5}} \\
\cmidrule(lr){3-4}  \cmidrule(lr){5-6}  
\textbf{Model} & \textbf{Judge} & \textbf{Macro}~$(\uparrow)$  & \textbf{Micro}~$(\uparrow)$ & \textbf{Macro}~$(\uparrow)$ & \textbf{Micro}~$(\uparrow)$ & \textbf{Avg.}~$(\uparrow)$ \\ \midrule 
GPT-4V &	&20.4	&35.5	&19.6	&25.8	&25.3 \\
MAIRA-1	& &38.6	&55.7	&47.7	&56.0	&49.5 \\
MAIRA-2	& &41.6	&58.1	&50.4	&59.1	&52.3 \\
Med-PaLM M  (12B)& 	&37.3	&51.4	&50.6	&56.5	&49.0 \\
Med-PaLM M (84B)&	&39.8	&53.6	&51.6	&57.9	&50.7 \\
% Med-PaLM M (562B)&	&37.3	&51.4	&50.6	&56.5	&49.0 \\
Med-PaLM M (562B)& &37.8	&51.6	&49.9	&56.3	&48.9 \\
LLaVA-Rad&	&39.5	&57.3	&47.7	&57.4	&50.5 \\   \midrule \midrule %\cmidrule(lr){3-7}  
CheXagent	&&38.9		&50.9		&47.6		&54.1		&47.9	 \\ \midrule 
+DPO	&&41.5$^*$	(6.87)	&54.1$^*$	(6.18)	&51.8$^*$	(8.83)	&58.3$^*$	(7.78)	&51.4$^*$	(7.43) \\
+LC-DPO	&&37.6	(-3.32)	&52.1	(2.20)	&47.5	(-0.02)	&55.2$^*$	(1.92)	&48.1	(0.45) \\
+IPO	&GREEN&39.0	(0.39)	&52.9$^*$	(3.87)	&48.9	(2.88)	&56.5$^*$	(4.36)	&49.3	(3.06) \\
+KTO	&&40.7	(4.81)	&55.0$^*$	(8.07)	&51.6$^*$	(8.52)	&59.3$^*$	(9.54)	&51.7$^*$	(7.93) \\
+ORPO	&&41.4$^*$	(6.64)	&55.0$^*$	(8.03)	&51.5$^*$	(8.20)	&58.3$^*$	(7.69)	&51.5$^*$	(7.69) \\ \midrule 
+DPO	&&42.5$^*$	(9.37)	&56.1$^*$	(10.1)	&54.7$^*$	(15.0)	&61.6$^*$	(13.8)	&53.7$^*$	(12.2) \\
+LC-DPO	&&42.3$^*$	(8.75)	&56.3$^*$	(10.6)	&52.7$^*$	(10.8)	&60.8$^*$	(12.4)	&53.0$^*$	(10.8) \\
+IPO	&BERTScore&43.1$^*$	(10.8)	&57.0$^*$	(11.9)	&53.6$^*$	(12.6)	&61.5$^*$	(13.6)	&53.8$^*$	(12.4) \\
+KTO	&&44.0$^*$	(13.3)	&58.0$^*$	(13.9)	&54.0$^*$	(13.5)	&62.3$^*$	(15.1)	&54.6$^*$	(14.0) \\
+ORPO	&&42.4$^*$	(9.16)	&56.7$^*$	(11.4)	&52.5$^*$	(10.5)	&60.7$^*$	(12.2)	&53.1$^*$	(10.9) \\ \midrule 
CheXagent-2	&&44.6		&57.8		&55.5		&62.4		&55.1	\\ \midrule 
+DPO	&GREEN&\textbf{45.8	(2.6)}	&\textbf{59.7$^*$	(3.3)}	&\textbf{56.0	(0.8)}	&\textbf{64.1	(2.7)}	&\textbf{56.4	(2.4)} \\
+DPO	&BERTScore&43.0	(-3.6)	&\textbf{59.7$^*$	(3.3)}&53.5	(-3.6)	&63.8	(2.3)	&55.0	(-0.1) \\
%& &  \multicolumn{5}{c}{\textbf{BERTScore-as-Judge}} \\  \cmidrule(lr){3-7}  
% CheXagent	&  \multicolumn{5}{c}{\textbf{BERTScore-as-Judge}} \\     \cmidrule(lr){2-6}  
\bottomrule
\end{tabular}
}
\caption{CheXbert scores on the MIMIC-CXR test set. Percentage change compared to SFT baseline in brackets. Best results in bold. $^*$statistically significantly different from SFT baseline at a 10\% level based on confidence intervals in Table ~\ref{tab:appendix_chexbert_ci}.}
\label{tab:chexbert}
\end{table*}

To further emphasize this issue, we show the best configurations according to GREEN, and LC-GREEN, in Table \ref{tab:avg_length}. For CheXagent, using GREEN to guide the hyperparameter search leads to substantially more verbose reports for DPO, LC-DPO ($\alpha=0.001$), IPO, and KTO. Qualitative evaluation of the resulting reports indicate that the added verbosity was mostly due to exact, semantically or syntactically, repetitions. This is very likely a manifestation of reward hacking via length exploitation. To mitigate this issue, we use LC-GREEN instead of GREEN to guide the hyperparameter search.\footnote{We opt for this approach, in lieu of obtaining new preference data and re-running the alignment due to computational constraints.} As shown in Table~\ref{tab:avg_length}, this leads to substantially less added verbosity. For CheXagent-2, it is less clear-cut whether length exploitation is present, as the highest and lowest scores are obtained with a similar verbosity. Note that this is in spite of a much more significant spread in average length of the reports in the chosen and rejected subsets. One possible explanation is that the SFT baseline produces substantially shorter reports than the references, and the increase in verbosity actually pushes the average closer to that of the references, effectively closing the gap rather than extending it by overshooting the average length of the references. 

\subsection{Judge Optimization Results}
Results, using single runs, for GREEN, LC-GREEN, and the BERTScore on the MIMIC-CXR and CheXpert Plus test sets are available in Table~\ref{tab:nxm}. Additional results for ROUGE-L and BLEU-4 are available in Table~\ref{tab:nxm_appendix}. As expected, using GREEN as Judge results in the largest boost in GREEN, and using the BERTScore in the BERTScore. In particular, the top performing configuration on the MIMIC-CXR test set for CheXagent according to GREEN and LC-GREEN is KTO, using GREEN as Judge, boosting these metrics by 31.9\% and 34.1\% percent, respectively. For CheXagent, the top performing configuration according to the BERTScore is obtained by KTO, using the BERTScore as Judge. We observe similar trends for CheXagent-2 with DPO yielding an increase of 18.9\% and 14.1\% for GREEN and LC-GREEN, respectively when using GREEN as Judge. We observed overall similar trends for CheXbert Plus, despite representing two different distributions: MIMIC-CXR was collected in an emergency department and CheXpert Plus was collected from in- and out-patient centers.

\subsection{Generalization to CheXbert Scores}
\begin{table*}[!htbp]
\centering
\resizebox{0.8\linewidth}{!}{%
\begin{tabular}{@{}lcccccccc@{}}
\toprule
& \multicolumn{8}{c}{\textbf{Accuracy}~$(\uparrow)$} \\ \cmidrule(lr){3-9}  
\multirow{2}{*}{\textbf{Model}} &\multirow{2}{*}{\textbf{Judge}} & \multirow{2}{*}{\textbf{View Classification}} & \textbf{Binary Image}   & \textbf{Single Disease} & \textbf{Multi Disease}  & \textbf{Visual Question} & \textbf{Image-Text}   & \multirow{2}{*}{\textbf{Avg.}} \\
                                &         &                                      & \textbf{Classification} & \textbf{Identification} & \textbf{Identification} & \textbf{Answering}       & \textbf{Reasoning}    &                                \\ \midrule
CheXagent	&&	98.5		&83.8			&62.8			&69.0			&75.5		&66.3 &76.0\\ \midrule 
+DPO		&&98.5		&84.0			&63.2			&68.7			&75.5		&66.1 & 76.0\\
+LC-DPO		&&98.5		&83.7			&62.2			&68.3			&75.5		&65.3 & 75.6\\
+IPO		&GREEN&98.3		&83.8			&62.7			&68.3			&76.8		&65.5 & 75.9\\
+KTO		&&98.3		&83.8			&63.2			&68.7			&74.6		&65.0 & 75.6\\ 
+ORPO		&&98.2		&83.5			&63.5			&68.6			&74.3		&64.2 & 75.4\\ \midrule 
+DPO		&&98.5		&84.3			&62.8			&68.4			&75.1		&66.8 & 76.0\\
+LC-DPO		&&98.5		&84.1			&63.2			&68.3			&75.1		&65.5 & 75.8\\
+IPO		&BERTScore&98.5		&83.8			&62.9			&68.2			&76.0		&66.3 & 76.0\\
+KTO		&&98.5		&84.2			&63.2			&68.5			&76.3		&66.3 & 76.2\\ 
+ORPO		&&98.3		&84.1			&62.7			&68.8			&73.9		&66.3 & 75.7\\  \midrule 
CheXagent-2		&&99.0		&83.0			&65.5			&83.9			&83.2		&78.7		&82.2 \\ \midrule 
+DPO		&GREEN&99.2		&84.8			&66.7			&84.4			&80.5		&77.9		&82.2 \\
+DPO	&BERTScore&99.2		&83.0			&65.9			&84.5			&82.3		&78.2		&82.2 \\ 
                        
\bottomrule
                
\end{tabular}}
\caption{Performance on six image perception and reasoning tasks different from RRG on the datasets listed in \S\ref{dataset}.}
\label{tab:alignment_tax}
\end{table*}

Although GREEN is a high quality and clinically relevant metric, the observed performance gains might be inflated due to the fact that we used GREEN as Judge for preference data generation. Thus, we instead turn to the CheXbert scores to be our silver-standard, a low-cost approximation of expert human (radiologists) judgment. CheXbert scores are ubiquitous in this setting and we include results from a representative sample of recent state-of-the-art medical VLMs for the RRG tasks: GPT-4V\footnote{\url{https://openai.com/index/gpt-4v-system-card/}.}, MAIRA-1~\citep{hyland2023maira}, MAIRA-2~\citep{bannur2024maira2}, Med-PaLM~\citep{tu2024towards}, and LLaVA-Rad~\citep{chaves2024llavarad}. We report macro and micro averaged F1-scores for the full 14 categories as well as the 5 categories subset. In addition, we provide the average across these scores. Results for our method are averages based on 1000 bootstrap samples, with confidence intervals available in Table~\ref{tab:appendix_chexbert_ci}. 

For CheXagent, which is comparable to LLava-Rad and MAIRA-1 prior to alignment, our method boosts the average CheXbert scores by up to 7-8\% and 14\% using GREEN and the BERTScore, respectively. Very interestingly, and unexpectedly, the top performing setup for CheXagent using the BERTScore as Judge is better than that for GREEN as Judge. One possible explanation for this is reward overoptimziation, as this seems to be a more prominent issue for GREEN than for the BERTScore. 
For CheXagent-2, which is the overall state-of-the-art prior to alignment and second only to MAIRA-2\footnote{Notably, MAIRA-2 use additional information at train and test time, including the \emph{radiologist-written} prior report, when available.} in terms of micro F1-14 scores, we can see that it is possible to improve performance even further using our proposed method. We note, however, that only the micro F1-14 scores are statistically significantly different from CheXbert-2, at a 10\% level, for both GREEN and the BERTScore as Judge. Thus, we depict that it is possible to improve upon the already very strong micro averaged F1-scores, even when employing a general domain NLG metric like the BERTScore. Granular results, for each of the 14 categories, are available in Table~\ref{tab:chexbert_appenidx}. 

\subsection{Alignment Tax Analysis}
While RLHF is powerful, it has been observed that it might lead to performance degradations or, forgetting~\citep{askell2021general,ouyang2022training}.~\citet{ouyang2022training} assessed such an alignment tax by evaluating the aligned policies on several natural language processing (NLP) benchmarks. Inspired by this, we benchmark the SFT baseline and the aligned policies on six diverse image perception and reasoning tasks using datasets listed in \S\ref{dataset}. Although there are some minor variations, on average, the performance matches that of the SFT baselines. Thus, our method substantially improves the quality of generated reports without compromising the quality of other image understanding tasks. 

\subsection{Qualitative Analysis}
\begin{figure}
    \centering
    \includegraphics[width=\linewidth]{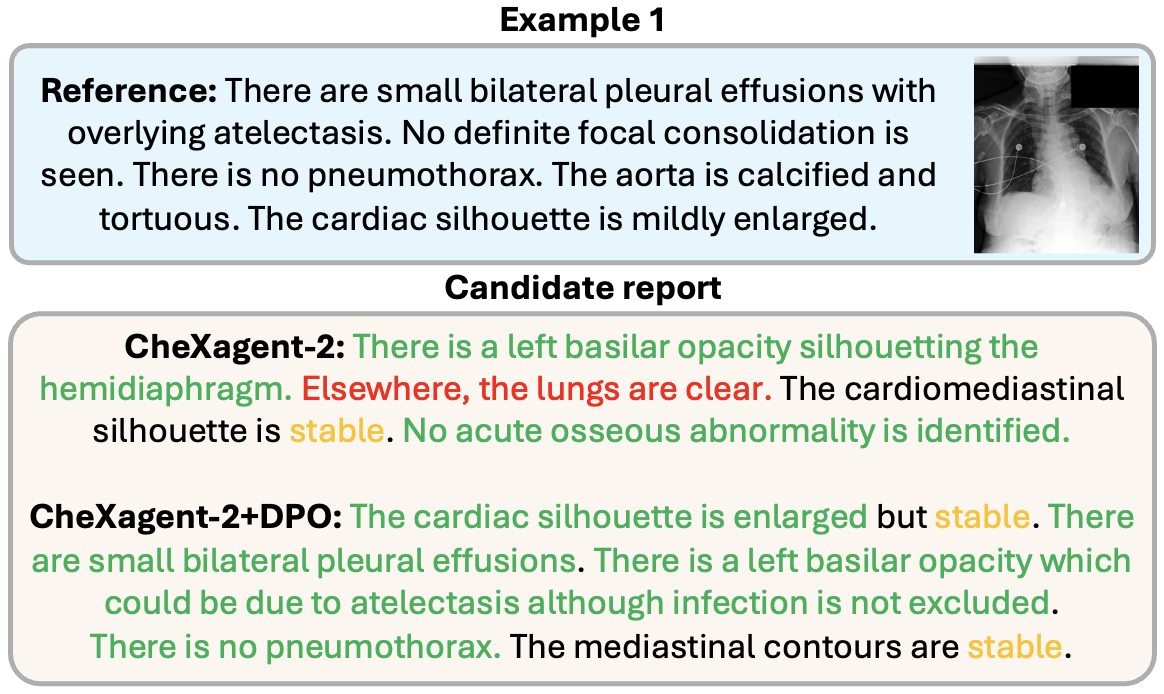}
    \caption{Color-coded candidate reports. Green and red represent correct and incorrect. Orange highlights references to a prior imaging study.}
    \label{fig:qualitative}
\end{figure}

To further shed light on policy behavior pre and post alignment, a qualitative study was conducted by selecting three interesting cases from the MIMIC-CXR test set, covering a range of positive and negative findings and the presence and absence of indications. A board-certified radiologist was then asked to color-code candidate reports from the SFT baselines and aligned policies. Results for CheXagent-2 using GREEN as Judge for the first example are available in Fig.~\ref{fig:qualitative}. For this case, despite being very strong, CheXagent-2 exhibited two errors: 1) it incorrectly reported clear lungs and 2) it failed to detect small bilateral pleural effusions. Notably, the aligned version addressed both of these errors. A similar conclusion is true for the other two cases. In addition, GREEN score as Judge seems to be superior to BERTScore and we observe exact, and semantically equivalent, repetitions in some cases. Full results, along with an extensive discussion, are available in \S\ref{appenidx:qualitative}.

\section{Conclusion}
Our study highlights the significant potential of including preference fine-tuning in the post-training pipeline of medical VLMs. By developing an automated pipeline we effectively address the prohibitive cost of obtaining radiologist preferences at scale. Using our approach, we have shown that DAAs can substantially improve AI-generated radiology reports in clinically meaningful ways \emph{without additional radiologist feedback}. Our approach achieves state-of-the-art performance on the MIMIC-CXR dataset while maintaining robust capabilities across diverse visual reasoning tasks. The surprising effectiveness of, inexpensive, general-domain NLG metrics for preference pair generation suggests a promising path forward for computationally efficient exploration of online alignment algorithms in this setting.

\section{Limitations}
Our work focuses on only two VLMs, CheXagent~\citep{chen2024chexagent} and CheXagent-2~\citep{chen2024chexagent2}. While different in terms of underlying architecture, information used, and baseline performance, it would be of interest to additionally study VLMs from other families and sizes. In addition, our study of CheXagent-2 was more limited in nature and should be considered more preliminary. 

Moreover, we treat clinically relevant metrics such as GREEN and CheXbert scores as the silver standard. While these metrics are highly relevant, a thorough evaluation involving clinical experts, radiologists, should be conducted for this study to be considered complete. Our qualitative analysis is a first step in this direction. However, conducting larger-scale reader studies remains an important direction for future work. 
We acknowledge potential biases beyond verbosity, such as societal biases related race, sex, or other demographic factors that may be embedded either in the underlying data or in the Judge. These potential biases warrant thorough investigation and mitigation in future work.

In addition, our hyperparameter search is non-exhaustive and it is possible that the relative ranking of the methods considered would change with a more extensive search. 

Finally, we restrict ourselves to only offline DAAs. This leaves out a range of very competitive alignment algorithms, including on-policy RL algorithms, as well as the online, or iterative, counterparts to the DAAs considered. The recent success of DeepSeek-R1~\citep{guo2025deepseekr1} has led to a resurgence of interest in on-policy RL alignment algorithms. In particular, developing ``verification functions'' for the RRG task and employing reinforcement learning with verifiable rewards (RLVR)~\citep{lambert2025tulu3} is an very interesting avenue for future work. 

\section{Acknowledgements}
We would like to acknowledge compute support from Microsoft, Google, and Stanford Marlowe~\citep{kaper2025marlowe}. D.H. is supported by MedTechLabs, the Göran Gustafsson foundation, and the foundation for Technical Scientific Research and Education. J.X. acknowledges support from the Canadian Institutes of Health Research (CIHR) and Oxford University Press. S.O. acknowledges funding from the DFG (German Research Foundation, ID: 517316550). M.V. is supported by graduate fellowship awards from the Department of Defense (NDSEG), the Knight-Hennessy Scholars program at Stanford University, and the Quad program. C.B. receives research support from the Promedica Foundation, Chur, CH. A.C. receives research support from NIH grants R01 HL167974, R01HL169345, R01 AR077604, R01 EB002524, R01 AR079431, P41 EB027060; ARPA-H grants AY2AX000045 and 1AYSAX0000024-01; and NIH contracts 75N92020C00008 and 75N92020C00021. Z.C. receives support from the AIMI-AWS Cloud Credit Program and the HAI-Google research grant.


\bibliography{custom}

%%%%%%%%%%%%%%%%%%%%%%%%%%%%%%%%%%%%%%%%%%%%%%%%%%%%%%%%%%%%%%%%%%%%%%%%%%%%%%%
%%%%%%%%%%%%%%%%%%%%%%%%%%%%%%%%%%%%%%%%%%%%%%%%%%%%%%%%%%%%%%%%%%%%%%%%%%%%%%%
% APPENDIX
%%%%%%%%%%%%%%%%%%%%%%%%%%%%%%%%%%%%%%%%%%%%%%%%%%%%%%%%%%%%%%%%%%%%%%%%%%%%%%%
%%%%%%%%%%%%%%%%%%%%%%%%%%%%%%%%%%%%%%%%%%%%%%%%%%%%%%%%%%%%%%%%%%%%%%%%%%%%%%%
\newpage
\appendix
\onecolumn
\section{RLHF and DAAs}
\label{appenidx:daas}
RLHF~\citep{ziegler2020finetuning,stiennon2020learning,ouyang2022training} is based on the constrained reward maximization objective 
\begin{equation}
    \max_{\pi_{\theta}} \mathbb{E}_{x \sim \mathcal{D}, y\sim \pi_{\theta}(y|x)}[R_\psi (x, y)]-\beta \mathbb{D}_{\text{KL}}\left[\pi_{\theta}(y|x) || \pi_{\text{ref}} (y|x)\right],
    \label{eq:rlhf_obj}
\end{equation}
where $\mathbb{D}_{\text{KL}}$ is the Kullback-Leibler (KL) divergence and $\pi_{\text{ref}}$ is the reference policy. $R_\psi$ is the proxy reward model learned on a dataset of human preferences $\mathcal{D}=\{x^{(n)}, y_c^{(n)}, y_r^{(n)}\}_{n=1}^N,$ where $y_c$ and $y_r$ denote the chosen and rejected completions for the prompt $x$, such that $y_c \succ y_r | x.$ 

Whilst extremely powerful, RLHF is computationally heavy, involves several steps, and can be tricky to implement in practice. Relatively recently, a new class of algorithms called DAAs~\citep{rafailov2024scaling} have become increasingly popular.\footnote{In this paper, we use this terminology more loosely than in ~\citet{rafailov2024scaling}.} This class of algorithms re-parameterize the reward model via a change-of-variables using the closed-formed solution to the objective in \eqref{eq:rlhf_obj}, effectively bypassing both the reward modeling and reinforcement learning (RL) stages. Resulting in algorithms that remain performant yet computationally more light weight and easier to implement. DPO~\citep{rafailov2023direct} was the first in this category and remains one of the most popular versions.

DPO exploits the closed-formed solution to \eqref{eq:rlhf_obj},  $\pi(y|x)\propto \pi_{\text{ref}}(y|x)\exp(R(x,y)/\beta)$ and the 
Bradley-Terry (BT) model~\citep{bradley1952rank} of human preferences $p^*(y_1 \succ y_2 | x) = \sigma(\exp(R^*(x,y_1))-\exp(R^*(x,y_2))),$
where $R^*$ is the latent reward model, $\exp$ is the exponential function, and $\sigma$ is the logistic function. The reward can be isolated and written as a function of the policy $R(x,y)=\beta \log \frac{\pi(y|x)}{\pi_{\text{ref}}(y|x)}.$ This re-parametrization can be applied to the latent reward $R^*$ and substituted into the BT model, 
$p^*(y_1 \succ y_1 | x) = \sigma \left(\beta \log \frac{\pi^*(y_1|x)}{\pi_{\text{ref}}(y_1|x)} - \beta \log \frac{\pi^*(y_2|x)}{\pi_{\text{ref}}(y_2|x)}\right),$ where $\pi^*$ is the optimal policy corresponding to the latent reward. Crucially, the probability of human preferences is now in terms of the policy instead of the reward model. A parameterized policy $\pi_\theta$ can then be learned via a simple classification loss over the preference data 
\begin{equation*}
    \mathcal{L}_{\text{DPO}}(\theta) =  
    -\log \sigma\left(\beta \log \frac{\pi_\theta\left(y_c \mid x\right)}{\pi_{\mathrm{ref}}\left(y_c \mid x\right)}-
    \beta \log \frac{\pi_\theta\left(y_r \mid x\right)}{\pi_{\mathrm{ref}}\left(y_r \mid x\right)}\right). 
\end{equation*}
Hence, this change-of-variables has transformed a loss over rewards into a loss over policies.

\section{Implementation Details}
\subsection{Data Details}
We use the official train, validation, and test splits for the MIMIC-CXR~\citep{johnson2019mimic} and CheXpert Plus\cite{chambon2024chexpertplus} datasets. 
A key difference between CheXagent and CheXagent-2 is that CheXagent-2 additionally use indications, which provide clinical context, to aid in the RRG task. This means that CheXagent-2 imposes a stricter data requirement (i.e. each image-report pair must have a corresponding indication). Due to this, CheXagent-2 has slightly fewer examples than CheXagent for the RRG task. The number of examples in each split is available in Table \ref{tab:split}. For MIMIC-CXR, the difference in test and validation is so small that it is negligible for the RRG task. On the other hand, for CheXpert Plus the difference is quite substantial. However, this is immaterial for any conclusions in this paper as we do not directly compare the models with each other on the CheXpert Plus data. 

The view classification, binary image classification, single disease identification, multi disease identification, VQA, and image-text reasoning test data from the RSNA~\citep{shih2019rsna}, SIIM~\citep{siim}, OpenI~\citep{demner2016openi}, SLAKE~\citep{liu2021slake}, and Rad-Restruct~\citep{pellegrini2023radrestruct} datasets are processed as in ~\citet{chen2024chexagent2}. 

\begin{table}[h]
    \centering
    \resizebox{0.6\linewidth}{!}{%
    \begin{tabular}{l*{6}{c}}
    \toprule
    & \multicolumn{3}{c}{\textbf{CheXagent}} & \multicolumn{3}{c}{\textbf{CheXagent-2}} \\
    \cmidrule(lr){2-4}  \cmidrule(lr){5-7} 
    \textbf{Dataset} & \textbf{Train} & \textbf{Validation} & \textbf{Test} & \textbf{Train} & \textbf{Validation} & \textbf{Test} \\ \midrule 
    MIMIC-CXR &148463& 1164& 2309& 148090& 1162& 2300 \\
    CheXpert Plus & & &208 & & & 151 \\ \bottomrule 
    \end{tabular}
    }
    \caption{Number of examples in each split of MIMIC-CXR and CheXpert plus for CheXagent and CheXagnet-2.}
    \label{tab:split}
\end{table}

\subsection{Training Details}
\label{appendix:implementation}
All models are trained using a machine with 4xA100, 4xA6000 or 8xH100 GPUs using learning rate learning rate $10^{-6}$. We set global batch size to 32 for CheXagent~\citep{chen2024chexagent} and 64 for CheXagent-2~\citep{chen2024chexagent2}. Each model is trained for one epoch. The image encoder is frozen while we train the LLM. Due to compute constraints, we only tune hyperparameters that are specific of the DAAs considered while keeping everything else fixed. An overview is given in Table \ref{tab:hyperparameters}, including optimal configurations. This non-exhaustive search was based on previous work and our initial experiments. For CheXagent, each $\lambda \in [0.5, 1.0, 4.0, 5.0]$, ORPO resulted in a model which produced a special token at odd places, leading to a crash of our evaluation pipeline. We address this by catching the error and set the special token to the padding token. The same issue emerged for some $\lambda$ for CheXagent-2.  
\begin{table}[h]
    \centering
    \resizebox{\linewidth}{!}{%
    \begin{tabular}{llc}
    \toprule
    \textbf{Algorithm} & \textbf{Objective} & \textbf{Hyperparameters} \\ \midrule
    DPO~\citep{rafailov2023direct} & $-\log \sigma\left(\beta \log \frac{\pi_\theta\left(y_c \mid x\right)}{\pi_{\mathrm{ref}}\left(y_c \mid x\right)}-\beta \log \frac{\pi_\theta\left(y_r \mid x\right)}{\pi_{\mathrm{ref}}\left(y_r \mid x\right)}\right)$ & $\beta \in [0.01,0.05^{*,\dagger},0.1]$  \\ 
    LC-DPO~\citep{park2024disentangling} & $-\log \sigma\left(\beta \log \frac{\pi_\theta\left(y_c \mid x\right)}{\pi_{\mathrm{ref}}\left(y_c \mid x\right)}-\beta \log \frac{\pi_\theta\left(y_r \mid x\right)}{\pi_{\mathrm{ref}}\left(y_r \mid x\right)}+\alpha(|y_c|-|y_r|)\right)$ & $\beta \in [0.01,0.05^{*,\dagger},0.1], \alpha \in [0.001, 0.01^{*,\dagger}]$  \\ 
     IPO~\citep{azar2023general} & $ \left( \log \frac{\pi_\theta(y_c|x)}{\pi_{\text{ref}}(y_c|x)} - \log \frac{\pi_\theta(y_r|x)}{\pi_{\text{ref}}(y_r|x)} - \frac{1}{2\tau} \right)^2 $ & $\tau \in [0.1,0.5,1.0^{*,\dagger}]$ \\ 
     KTO~\citep{ethayarajh2024kto} & $-\lambda_c \sigma \left( \beta \log \frac{\pi_\theta(y_c|x)}{\pi_{\text{ref}}(y_c|x)} - z_{\text{ref}} \right) +  \lambda_r \sigma \left( z_{\text{ref}} - \beta \log \frac{\pi_\theta(y_r|x)}{\pi_{\text{ref}}(y_r|x)} \right)$ & $\beta \in [0.01,0.05^{*},0.1^{\dagger}], \lambda_c=\lambda_r$ \\ 
     & $\text{where} \,\, z_{\text{ref}}=\mathbb{E}_{(x,y)\sim \mathcal{D}}[\beta  \mathbb{D}_{\text{KL}} \left(\pi_{\theta}(y|x)||\pi_{\text{ref}}(y|x)\right)]$ &  \\ 
     ORPO~\citep{hong2024orpo} & $-\log p_\theta(y_c|x) - \lambda  \log \sigma \left(\log \frac{p_\theta(y_c|x)}{1 - p_\theta(y_c|x)} - \log \frac{p_\theta(y_r|x)}{1 - p_\theta(y_r|x)}  \right),$ & $\lambda \in [0.5, 1.0, 4.0^{\dagger}, 5.0^*]$ \\ 
     & $\text{where} \,\, p_\theta(y|x) = \exp\left( \frac{1}{|y|} \log \pi_\theta(y|x) \right)$ &  \\ \bottomrule
    \end{tabular}
    }
    \caption{Hyperparameter search for all direct alignment algorithms (DAAs) considered in this paper. For CheXagent, $^*$ and $^{\dagger}$ denotes best for GREEN and the BERTScore as Judge, respectively. For CheXagent-2, $\beta=0.1$ was found optimal for both GREEN and the BERTScore.}
    \label{tab:hyperparameters}
\end{table}

\subsection{Sampling Details}
In this paper, we set up CheXagent to treat the cases where the two CXRs are a frontal and a lateral image (i.e. from the same point in time) and a frontal and a prior frontal image (i.e. from two points in time) as separate cases with a separate prompt. This is in contrast to CheXagent-2, which employs the same prompt for both cases. Moreover, we employed stochastic sampling with temperature $1.0$ for CheXagent at test time, while CheXagent-2 is based on greedy sampling, both with beam size set to $1$. For preference pairs generation, both employ stochastic sampling with temperature $1.0$ and beam size set to $1.$ These differences will explain parts of the performance differences between CheXagent and CheXagent-2. However, this is of minor concern to this study as we are mainly interested in the effect of preference fine-tuning, comparing the same setup prior and post alignment.  

\subsection{Evaluation Details}
GREEN~\citep{ostmeier2024green} and GREEN-LC are based on the official implementation\footnote{\url{https://github.com/Stanford-AIMI/GREEN}.} using StanfordAIMI/GREEN-radllama2-7b. The BERTScore~\citep{zhang2019bertscore} used is from evaluate (v0.4.0) with distilbert-base-uncased as BERT model. The F1 CheXbert~\citep{smit2020combining} scores use f1chexbert (v0.0.2) with default configurations. ROUGE~\citep{lin2004rouge} is from rouge-score (v0.1.2) using rougeL. Finally, the BLEU~\citep{papineni2002bleu} is BLEU-4, based on a custom code, available in the code base corresponding to this project. 

CheXbert scores for GPT-4V\footnote{\url{https://openai.com/index/gpt-4v-system-card/}.}, MAIRA-1~\citep{hyland2023maira}, Med-PaLM~\citep{tu2024towards}, and LLaVA-Rad~\citep{chaves2024llavarad} are borrowed directly from Supplementary Table 1 in \citet{chaves2024llavarad}. The CheXbert scores for MAIRA-2~\citep{bannur2024maira2} are from Table D.1 in~\citet{bannur2024maira2}. 

\section{Additional results}
\subsection{Preference data}
\label{appenidx:preference}
\begin{figure}
    \centering
    \includegraphics[width=\linewidth]{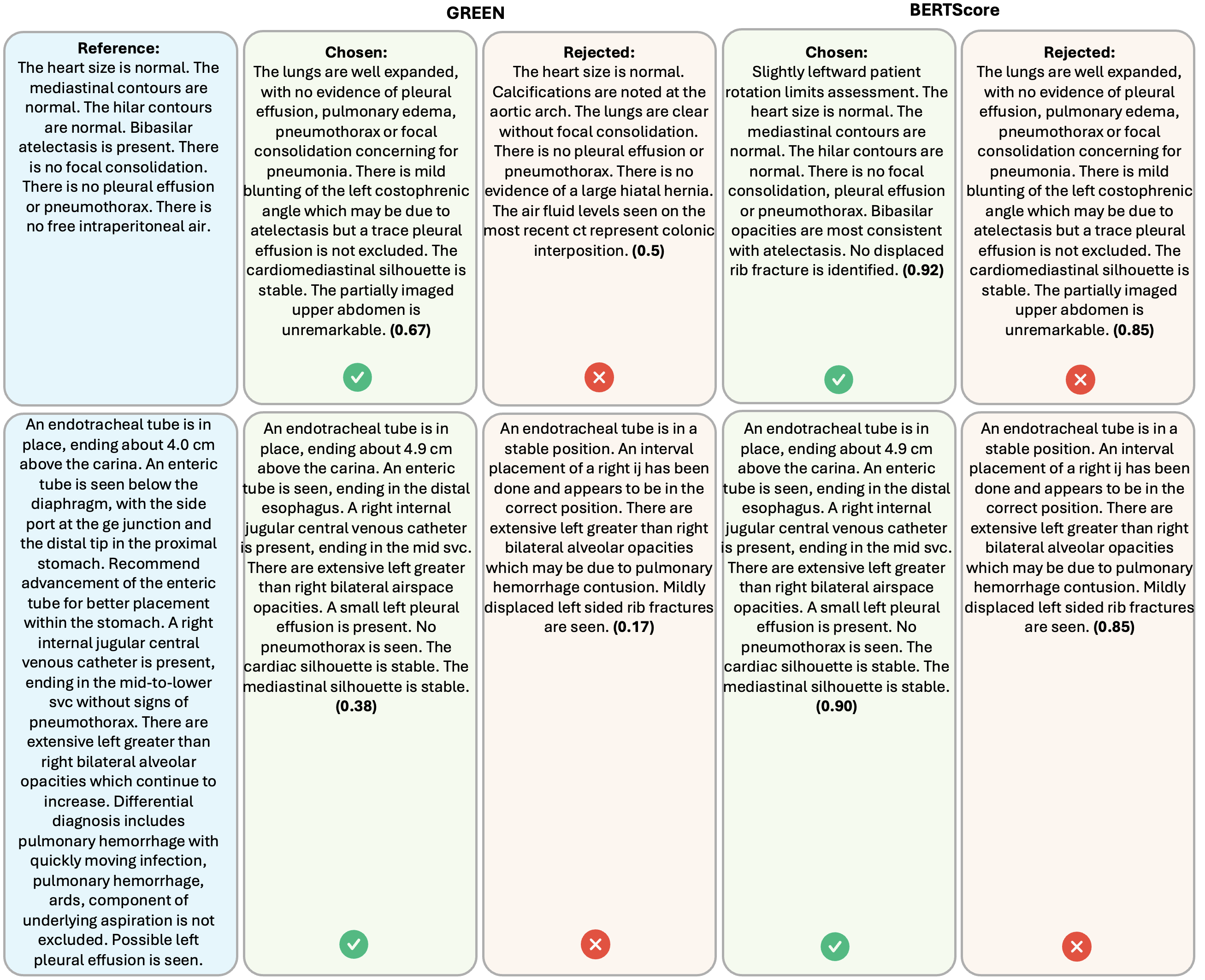}
    \caption{First two examples of chosen and rejected candidates for CheXagent in the MIMIC-CXR train set. Number in brackets is assigned score (GREEN or BERTScore).}
    \label{fig:chosen_rejected_chexagent}
\end{figure}

\begin{figure}
    \centering
    \includegraphics[width=\linewidth]{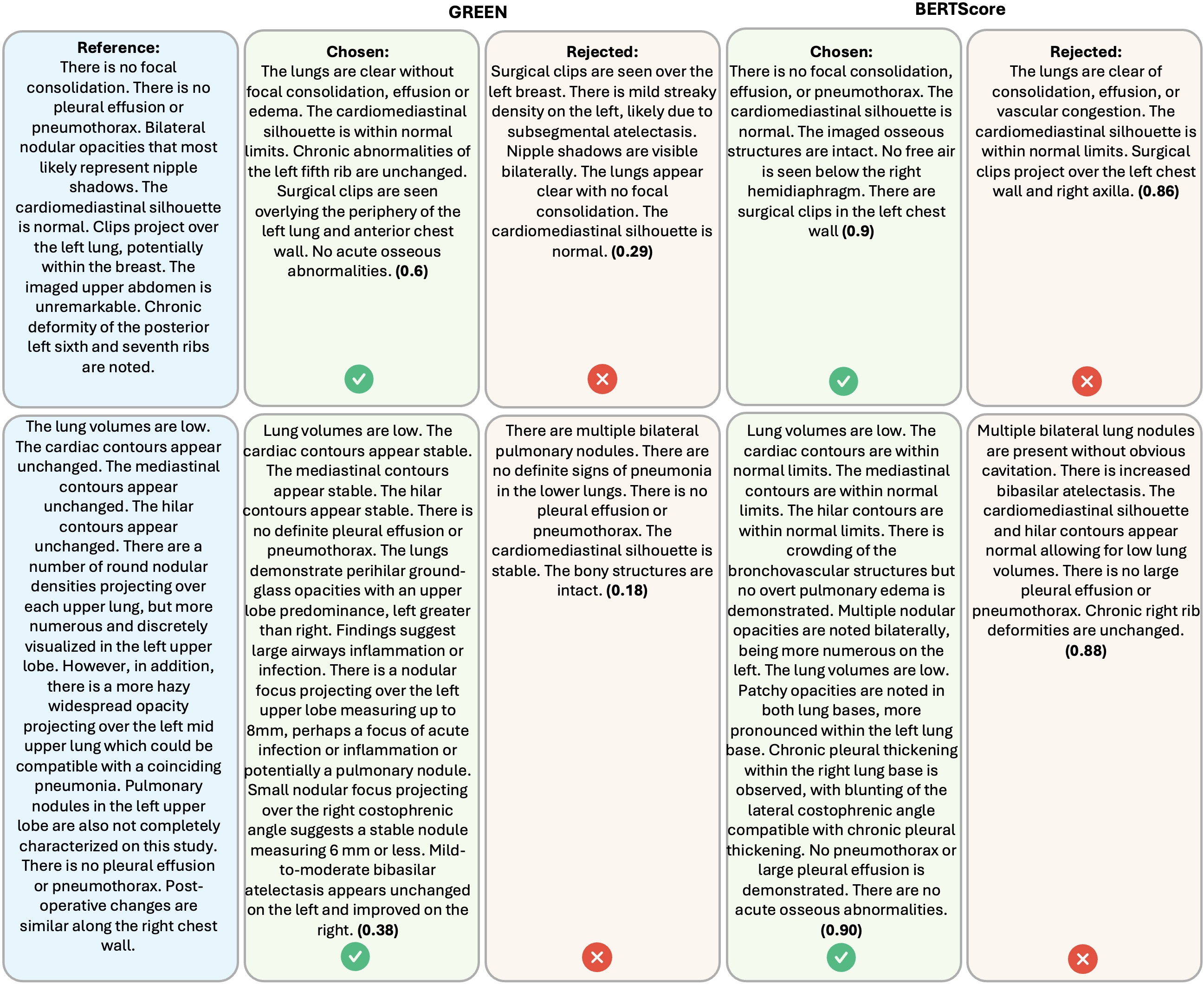}
    \caption{First two examples of chosen and rejected candidates for CheXagent-2 in the MIMIC-CXR train set. Number in brackets is assigned score (GREEN or BERTScore).}
    \label{fig:chosen_rejected_chexagent2}
\end{figure}

Summary statistics for the chosen and rejected subsets are available in Table \ref{tab:reward_summary}. We also report summary statistics of the length (in words) of the generated reports. Notably, the difference, or spread, in average length of the chosen and rejected subsets is slightly more pronounced for GREEN than for the BERTScore, 6.9\% compared to 5.8\% and much more significant overall for CheXagent-2 with a difference of 19.3\% and 17.1\%, respectively. To build further build intuition on the resulting chosen and rejected subsets we report CheXbert scores in Table~\ref{tab:preference_chexbert}. Scores for CheXagent and CheXagent-2 are not directly comparable since the training data is different. As expected, these scores indicate that the chosen subset is considerably better than the rejected. The average CheXbert scores for the chosen subset are more or less the same for GREEN and BERTScore as Judge. Similarly is true for the rejected subset, though now BERTScore as Judge results in slightly lower scores--meaning that the spread is marginally larger. Finally, as a qualitative study, we include a couple of examples of rejected and chosen candidates for CheXagent and CheXagent-2 is Fig.~\ref{fig:chosen_rejected_chexagent} and Fig.~\ref{fig:chosen_rejected_chexagent2}, respectively. We include the score assigned by the respective Judge in brackets.  

\begin{table*}[!htbp]
\centering
\resizebox{0.8\linewidth}{!}{%
\begin{tabular}{l*{12}{c}}
\toprule
& \multicolumn{6}{c}{\textbf{GREEN}} & \multicolumn{6}{c}{\textbf{BERTScore}} \\ \cmidrule(lr){2-7} \cmidrule(lr){8-13} 
& \multicolumn{3}{c}{\textbf{Metric}} &  \multicolumn{3}{c}{\textbf{Report Length}} & \multicolumn{3}{c}{\textbf{Metric}} &  \multicolumn{3}{c}{\textbf{Report Length}} \\
\cmidrule(lr){2-4} \cmidrule(lr){5-7} \cmidrule(lr){8-10}  \cmidrule(lr){11-13} 
\textbf{CheXagent} & \textbf{Mean} & \textbf{Median} & \textbf{Std.} & \textbf{Mean} & \textbf{Median} & \textbf{Std.} & \textbf{Mean} & \textbf{Median} & \textbf{Std.} & \textbf{Mean} & \textbf{Median} & \textbf{Std.} \\ 
\cmidrule(lr){1-1} \cmidrule(lr){2-7} \cmidrule(lr){8-13}
Chosen	&0.63	&0.60	&0.24 &56.3	&54.0	&20.2	&0.90	&0.90 &0.03	& 55.1 &53.0	&20.2\\
Rejected	&0.26	&0.22	&0.19	&52.7	&51.0	&24.9 & 0.85	&0.86	&0.04	&52.1	&50.0	&25.3 \\  \midrule 
\textbf{CheXagent-2} & \textbf{Mean} & \textbf{Median} & \textbf{Std.} & \textbf{Mean} & \textbf{Median} & \textbf{Std.} & \textbf{Mean} & \textbf{Median} & \textbf{Std.} & \textbf{Mean} & \textbf{Median} & \textbf{Std.} \\ 
\cmidrule(lr){1-1} \cmidrule(lr){2-7} \cmidrule(lr){8-13}
% Chosen	&0.57	&0.50	&0.25	&0.24	&0.20	&0.20 &0.91	&0.90	&0.03	&0.85	&0.86	&0.05 \\
% Rejected	&54.4	&52.0	&19.3	&45.6	&44.0	&22.8 &52.7	&51.0	&18.6	&45.0	&43.0	&23.8 \\
Chosen &0.57	&0.50	&0.25	&54.4	&52.0	&19.3	&0.91	&0.90	&0.03	&52.7	&51.0	&18.6 \\
Rejected &0.24	&0.20	&0.20	&45.6	&44.0	&22.8	&0.85	&0.86	&0.05	&45.0	&43.0	&23.8 \\
\bottomrule
\end{tabular}
}
\caption{Summary statistics of reference-based metric and report length in the chosen and rejected subsets using GREEN and the BERTScore as Judges.}
\label{tab:reward_summary}
\end{table*}

\begin{table*}[!htbp]
\centering
\resizebox{\linewidth}{!}{%
\begin{tabular}{l*{14}{c}}
\toprule
% & \multicolumn{6}{c}{\textbf{Chosen}} & \multicolumn{6}{c}{\textbf{Rejected}} \\ \cmidrule(lr){2-7} \cmidrule(lr){8-13} 

& & & \multicolumn{5}{c}{\textbf{Chosen}} &  \multicolumn{5}{c}{\textbf{Rejected}} \\
\cmidrule(lr){4-8} \cmidrule(lr){9-13}  
& & & \multicolumn{2}{c}{\textbf{F1-14}} & \multicolumn{2}{c}{\textbf{F1-5}} & & \multicolumn{2}{c}{\textbf{F1-14}} & \multicolumn{2}{c}{\textbf{F1-5}} \\
\cmidrule(lr){4-5} \cmidrule(lr){6-7} \cmidrule(lr){9-10}  \cmidrule(lr){11-12} 
\textbf{Model} & & \textbf{Judge }& \textbf{Macro}~$(\uparrow)$ & \textbf{Micro}~$(\uparrow)$ & \textbf{Macro}~$(\uparrow)$ & \textbf{Micro}~$(\uparrow)$ & \textbf{Avg.}~$(\uparrow)$ & \textbf{Macro}~$(\uparrow)$ & \textbf{Micro}~$(\uparrow)$ & \textbf{Macro}~$(\uparrow)$ & \textbf{Micro}~$(\uparrow)$ & \textbf{Avg.} ~$(\uparrow)$ \\ \midrule

\multirow{2}{*}{CheXagent}	&&GREEN	&0.574	&0.668	&0.649	&0.700	&0.648	&0.411	&0.506	&0.455	&0.507	&0.463 \\
 &&BERTScore	&0.582	&0.680	&0.647	&0.700	&0.652	&0.395	&0.488	&0.444	&0.493	&0.419 \\
\multirow{2}{*}{CheXagent-2}	&&GREEN	&0.488	&0.587	&0.550	&0.621	&0.561	&0.336	&0.416	&0.365	&0.415	&0.373 \\
&&BERTScore	&0.488	&0.594	&0.541	&0.615	&0.560	&0.316	&0.394	&0.346	&0.393	&0.362 \\

\bottomrule
\end{tabular}
}
\caption{CheXbert scores on the MIMIC-CXR on the chosen and rejected subsets.}
\label{tab:preference_chexbert}
\end{table*}

\newpage 
\subsection{Judge Optimization Results}
\begin{table*}[h]
\centering
\resizebox{0.9\linewidth}{!}{%
\begin{tabular}{l*{7}{c}}
\toprule
& & \multicolumn{5}{c}{\textbf{MIMIC-CXR}} \\
\cmidrule(lr){3-7}  
\textbf{Model} & \textbf{Judge} & \textbf{GREEN}~$(\uparrow)$  & \textbf{GREEN-LC}~$(\uparrow)$ & \textbf{BERTScore}$(\uparrow)$ & \textbf{BLEU-4}~$(\uparrow)$ & \textbf{ROUGE-L}~$(\uparrow)$ \\ \midrule 
CheXagent	&	&0.249		&0.218	&	0.856	&	0.041	&	0.274	\\ \midrule 
+DPO	& & 0.325	(30.6)	& 0.263	(20.5)	&0.862	(0.65)	&0.057	(39.9)	&0.293	(6.94) \\
+LC-DPO	& & 0.320	(28.7)	& 0.288	(32.0)	&0.864	(0.90)	&0.056	(38.4)	&0.302	(10.4) \\
+IPO	& GREEN & 0.326	(30.9)	& 0.282	(29.3)	&0.863	(0.84)	&0.059	(45.0)	&0.297	(8.57) \\
+KTO	& & \textbf{0.328	(31.9)}	& \textbf{0.293	(34.1)}	&0.867	(1.27)	&0.056	(38.8)	&0.305	(11.4) \\
+ORPO	& & 0.322	(29.4)	& 0.275	(26.2)	&0.862	(0.69)	&0.053	(29.5)	&0.290	(6.08) \\ \midrule % \cmidrule(lr){3-7}  
+DPO	& & 0.285	(14.4)	& 0.242	(11.0)	&0.869	(1.56)	&0.059	(44.2)	&0.309	(12.9) \\
+LC-DPO	& & 0.283	(13.5)	& 0.259	(18.7)	&0.871	(1.75)	&0.057	(41.2)	&\textbf{0.315	(15.1)} \\
+IPO	& BERTScore & 0.283	(13.6)	& 0.247	(13.1)	&0.870	(1.62)	&\textbf{0.060	(47.2)}	&0.309	(12.9) \\
+KTO	& & 0.304	(21.9)	& 0.279	(28.2)	&\textbf{0.872	(1.88)}	&0.056	(37.7)	&0.310	(13.4) \\
+ORPO	& & 0.291	(16.9)	& 0.265	(21.4)	&0.869	(1.57)	&0.054	(33.8)	&0.307	(12.0) \\ \midrule 
CheXagent-2	&	&	0.326		&0.297		&0.888		&0.136 	&0.350	\\ \midrule 
+DPO	&GREEN	&\textbf{0.387	(18.9)}	&\textbf{0.339	(14.1)}	&0.891	(0.301)	&0.150	(10.3)	&0.357	(1.93) \\
+DPO	&BERTScore	&0.352	(8.11)	&0.326	(9.58)	&\textbf{0.896	(0.949)}	&\textbf{0.153	(12.5)}	&\textbf{0.372	(6.37)} \\
& & & & & & \\
& & \multicolumn{5}{c}{\textbf{CheXpert Plus}} \\
\cmidrule(lr){3-7}  
\textbf{Model} & \textbf{Judge} & \textbf{GREEN}~$(\uparrow)$  & \textbf{GREEN-LC}~$(\uparrow)$ & \textbf{BERTScore}$(\uparrow)$ & \textbf{BLEU-4}~$(\uparrow)$ & \textbf{ROUGE-L}~$(\uparrow)$ \\ \midrule 
CheXagent &&	0.248	&	0.202	 &0.851	&	0.038	&	0.274	\\ \midrule 
+DPO	& &0.322	(29.9)	&0.222	(9.69)	&0.859	(0.91)	&0.053	(40.2)	&0.289	(5.25) \\
+LC-DPO	& &0.330	(32.8)	&\textbf{0.268	(32.3)}	&0.861	(1.26)	&0.049	(30.2)	&0.305	(10.6) \\
+IPO	& GREEN &0.334	(34.6)	&0.267	(31.9)	&0.861	(1.15)	&0.052	(39.1)	&0.295	(7.22) \\
+KTO	& &\textbf{0.341	(37.2)}	&0.266	(31.4)	&0.863	(1.42)	&0.052	(36.9)	&0.305	(10.9) \\
+ORPO	& &0.326	(31.4)	&0.232	(14.6)	&0.856	(0.64)	&0.046	(21.9)	&0.287	(4.63) \\ \midrule % \cmidrule(lr){3-7}  
+DPO	& &0.278	(12.0)	&0.208	(2.84)	&0.867	(1.89)	&0.054	(43.6)	&0.307	(11.3) \\
+LC-DPO	& &0.299	(20.3)	&0.248	(22.5)	&\textbf{0.871	(2.34)}	&\textbf{0.056	(47.6)}	&\textbf{0.318	(15.1)} \\
+IPO	& BERTScore &0.282	(13.6)	&0.218	(7.88)	&0.866	(1.75)	&0.052	(37.6)	&0.309	(12.0) \\
+KTO	& &0.308	(24.0)	&0.249	(23.1)	&0.869	(2.11)	&0.054	(43.8)	&0.316	(14.6) \\
+ORPO	& &0.298	(19.9)	&0.240	(18.4)	&0.870	(2.21)	&0.049	(30.8)	&0.314	(14.0) \\ \midrule 
CheXagent-2-3b	& &	0.349	&0.304	&0.892	&0.123	&0.361	 \\ \midrule 
+DPO	&GREEN	&\textbf{0.387	(10.89)}	&\textbf{0.320	(5.34)}&0.888	(-0.37)	&0.126	(2.74)	&0.356	(-1.30) \\
+DPO	&BERTScore	&0.359	(2.88)	&0.310	(1.78)	&\textbf{0.893	(0.15)}	&\textbf{0.130	(2.86)}	&\textbf{0.357	(-0.91)} \\
\bottomrule
\end{tabular}
}
\caption{Results on the MIMIC-CXR and CheXpert Plus test sets (percentage change compared to SFT baseline in brackets).  Best results in bold, separate for CheXagent and CheXagent-2.}
\label{tab:nxm_appendix}
\end{table*}

\newpage
\subsection{Generalization to CheXbert Scores}
\begin{table*}[!htbp]
\centering
\resizebox{0.8\linewidth}{!}{%
\begin{tabular}{l*{6}{c}}
\toprule
&& \multicolumn{2}{c}{\textbf{F1-14}} & \multicolumn{2}{c}{\textbf{F1-5}} \\
\cmidrule(lr){3-4}  \cmidrule(lr){5-6}  
\textbf{Model} & \textbf{Judge} & \textbf{Macro}~$(\uparrow)$  & \textbf{Micro}~$(\uparrow)$ & \textbf{Macro}$(\uparrow)$ & \textbf{Micro}~$(\uparrow)$ & \textbf{Avg.}~$(\uparrow)$ \\ \midrule 
GPT-4V &	&20.4	&35.5	&19.6	&25.8	&25.3 \\
MAIRA-1	& &38.6	&55.7	&47.7	&56.0	&49.5 \\
MAIRA-2	& &41.6	&58.1	&50.4	&59.1	&52.3 \\
Med-PaLM M (12B)& 	&37.3	&51.4	&50.6	&56.5	&49.0 \\
Med-PaLM M(84B)&	&39.8	&53.6	&51.6	&57.9	&50.7 \\
% Med-PaLM M (562B)&	&37.3	&51.4	&50.6	&56.5	&49.0 \\
Med-PaLM M (562B)& &37.8	&51.6	&49.9	&56.3	&48.9 \\
LLaVA-Rad&	&39.5	&57.3	&47.7	&57.4	&50.5 \\   \midrule \midrule %\cmidrule(lr){3-7}  

CheXagent	&&38.9$_{[37.8,40.0]}$	&50.9$_{[50.0,51.8]}$	&47.6$_{[46.3,48.8]}$	&54.1$_{[52.9,55.3]}$	&47.9$_{[46.9,48.8]}$ \\ \midrule 
+DPO	&&41.5$_{[40.2,42.9}]$	&54.1$_{[53.2,54.9}]$	&51.8$_{[50.4,53.1]}$	&58.3$_{[57.1,59.5}]$ &51.4$_{[50.4,52.4]}$ \\
+LC-DPO	&&37.6$_{[36.3,38.9}]$	&52.1$_{[51.1,53.1]}$	&47.5$_{[46.1,49.0]}$	&55.2$_{[53.9,56.4]}$	&48.1$_{[47.1,49.2]}$ \\
+IPO	&&39.0$_{[37.7,40.3}]$	&52.9$_{[52.0,53.8]}$	&48.9$_{[47.5,50.4]}$	&56.5$_{[55.3,57.7]}$	&49.3$_{[48.3,50.3]}$ \\
+KTO	&&40.7$_{[39.5,41.9}]$	&55.0$_{[54.1,55.9]}$	&51.6$_{[50.3,52.9]}$	&59.3$_{[58.1,60.4]}$	&51.7$_{[50.7,52.6]}$ \\
+ORPO	&&41.4$_{[40.1,42.7}]$	&55.0$_{[54.2,55.9]}$	&51.5$_{[50.1,52.8]}$	&58.3$_{[57.1,59.5]}$	&51.5$_{[50.6,52.5]}$ \\ \midrule 
+DPO	&&42.5$_{[41.4,43.7}]$	&56.1$_{[55.2,56.9]}$	&54.7$_{[53.5,55.9]}$	&61.6$_{[60.5,62.6]}$	&53.7$_{[52.8,54.6]}$ \\
+LC-DPO	&&42.3$_{[41.0,43.4}]$	&56.3$_{[55.5,57.2]}$	&52.7$_{[51.3,54.0]}$	&60.8$_{[59.6,61.9]}$	&53.0$_{[52.1,54.0]}$ \\
+IPO	&&43.1$_{[41.9,44.2}]$	&57.0$_{[56.2,57.8]}$	&53.6$_{[52.3,54.8]}$	&61.5$_{[60.4,62.5]}$	&53.8$_{[52.9,54.6]}$ \\
+KTO	&&44.0$_{[42.9,45.3}]$	&58.0$_{[57.2,58.9]}$	&54.0$_{[52.8,55.2]}$	&62.3$_{[61.2,63.4]}$	&54.6$_{[53.7,55.5]}$ \\
+ORPO	&&42.4$_{[41.2,43.7}]$	&56.7$_{[55.9,57.6]}$	&52.5$_{[51.3,53.6]}$	&60.7$_{[59.7,61.7]}$	&53.1$_{[52.2,53.9]}$ \\ \midrule 
CheXagent-2	&&44.6$_{[43.3,45.9}]$	&57.8$_{[56.9,58.7]}$	&55.5$_{[54.2,57.1]}$	&62.4$_{[61.2,63.6]}$	&55.1$_{[54.1,56.1]}$ \\ \midrule 
+DPO	&&\textbf{45.8}$_{[44.5,47.1}]$	&\textbf{59.7}$_{[58.9,60.6]}$	&\textbf{56.0}$_{[54.6,57.5]}$	&\textbf{64.1}$_{[63.0,65.3]}$	&\textbf{56.4}$_{[55.4,	57.4]}$ \\
+DPO	&&43.0$_{[41.7,44.4}]$	&\textbf{59.7}$_{[58.8,60.6]}$	&53.5$_{[52.1,54.9]}$	&63.8$_{[62.6,65.0]}$	&55.0$_{[54.1,56.0]}$ \\
\bottomrule
\end{tabular}
}
\caption{CheXbert scores on the MIMIC-CXR test set. 90\% confidence interval obtained by 1000 bootstrap samples in subscripts. Best results in bold.}
\label{tab:appendix_chexbert_ci}
\end{table*}

\begin{table*}[!htbp]
\centering
\resizebox{\linewidth}{!}{%
\begin{tabular}{l*{15}{c}}
\toprule
& \multicolumn{15}{c}{\textbf{F1-scores}~$(\uparrow)$} \\ \cmidrule(lr){3-16} \\ 
\textbf{Model} & \textbf{Judge} &	\textbf{ECm.}	&\textbf{Cmgl.}	&\textbf{LOpac.}	&\textbf{LLes.}	&\textbf{Edema}      	&\textbf{Cnsl.}	&\textbf{Pna.}	&\textbf{Atel.}	&\textbf{PEff.}	&\textbf{Pmtx.}	&\textbf{POth.}	&\textbf{Frac.}	&\textbf{SuDev.}	&\textbf{NoF.} \\ \midrule 
CheXagent&	&0.347	&0.620	&0.461	&0.171	&0.493	&0.158	&0.227	&0.453	&0.655	&0.444	&0.092	&\textbf{0.240}	&0.787	&0.304\\ \midrule 
+DPO	&&0.372	&0.675	&0.444	&0.186	&0.525	&0.226	&0.231	&0.452	&0.710	&0.416	&0.178	&0.217	&0.822	&0.365\\
+LC-DPO	&&0.385	&0.666	&0.380	&0.178	&0.430	&0.211	&0.158	&0.404	&0.665	&0.407	&0.083	&0.129	&0.835	&0.339\\
+IPO	&GREEN &0.402	&0.686	&0.395	&0.160	&0.472	&0.222	&0.206	&0.405	&0.661	&0.483	&0.090	&0.138	&0.822	&0.321\\
+KTO	&&0.388	&0.682	&0.422	&0.202	&0.542	&0.192	&0.180	&0.444	&0.721	&0.552	&0.074	&0.118	&\textbf{0.841}	&0.349\\
+ORPO	&&0.348	&0.684	&0.479	&0.201	&0.492	&0.224	&\textbf{0.247}	&0.475	&0.698	&0.511	&0.072	&0.177	&0.835	&0.365\\ \midrule 
+DPO	&&0.373	&0.688	&0.467	&0.192	&\textbf{0.580}	&\textbf{0.236}	&0.141	&0.503	&0.728	&0.483	&0.157	&0.239	&0.831	&0.336\\
+LC-DPO	&&0.301	&0.692	&\textbf{0.506}	&0.208	&0.567	&0.175	&0.107	&0.471	&0.731	&0.567	&0.167	&0.225	&0.832	&0.376\\
+IPO	&BERTScore &0.386	&0.690	&0.500	&\textbf{0.218}	&0.559	&0.190	&0.111	&0.500	&\textbf{0.739}	&0.550	&\textbf{0.195}	&0.220	&0.836	&0.342\\
+KTO	&&\textbf{0.430}	&\textbf{0.701}	&0.487	&0.217	&0.577	&0.172	&0.201	&\textbf{0.514}	&0.735	&\textbf{0.584}	&0.166	&0.197	&0.830	&0.360\\
+ORPO	&&0.377	&0.690	&0.491	&0.148	&0.562	&0.157	&0.231	&0.481	&0.736	&0.497	&0.116	&0.218	&0.826	&\textbf{0.418}\\ \midrule 
CheXagent-2	&&0.324	&0.672	&0.539	&\textbf{0.244}	&0.606	&\textbf{0.235}	&0.284	&0.543	&0.720	&\textbf{0.527}	&0.085	&\textbf{0.330}	&0.815	&0.333 \\ \midrule 
+DPO	&&\textbf{0.450}	&0.696	&0.493	&0.218	&0.606	&0.214	&\textbf{0.325}	&\textbf{0.545}	&0.738	&0.526	&\textbf{0.157}	&0.255	&\textbf{0.856}	&0.343 \\
+DPO	&&0.358	&\textbf{0.703}	&\textbf{0.547}	&0.166	&\textbf{0.622}	&0.127	&0.066	&0.478	&\textbf{0.749}	&0.550	&0.143	&0.271	&0.853	&\textbf{0.396} \\

\bottomrule
\end{tabular}
}
\caption{F1 scores on the MIMIC-CXR test set using 14 categories from the CheXbert labeler~\citep{smit2020combining}: Enlarged Cardiomediastinum (ECm.), Cardiomegaly (Cmgl.),  Lung Opacity (LOpac.), Lung Lesion (LLes.), Edema, Consolidation (Cnsl.), Pneumonia (Pna.), Atelectasis (Atel.), Pleural Effusion (PEff.), Pneumothorax (Pmtx.), Pleural Other (POth.), Fracture (Frac.), Support Devices (SuDev.), no Findings (NoF.). Best results in bold, separate for CheXagent and CheXagent-2.}
\label{tab:chexbert_appenidx}
\end{table*}

\newpage
\subsection{Qualitative Analysis}
\label{appenidx:qualitative}
To build some further intuition of behavior pre and post alignment, we conduct a qualitative study in which a board-certified radiologist was asked to color-code candidate reports from CheXagent, CheXagent-2, as well as their top performing aligned versions for GREEN and BERTScore as Judge, respectively, in terms of cheXbert scores in Table~\ref{tab:chexbert}. We selected three interesting, and representative, examples from MIMIC-CXR test set, including examples with positive and negative findings, only negative findings, and with and without indications. These were subsequently given to a radiologist to color-code, using the reference reports as well as the CXRs in png format (i.e. exact measurements are not possible). Three cases with six reports each, resulted in a total of 18 reports to be processed. The reports are color-coded as green (correct), red (incorrect), and orange to indicate where an reference to prior report is made though none was presented at the time of inference. For instance, the statement that something is ``stable'' may or may not be true, depending on the prior imaging study.

\begin{figure}[!htbp]
    \centering
    \includegraphics[width=\linewidth]{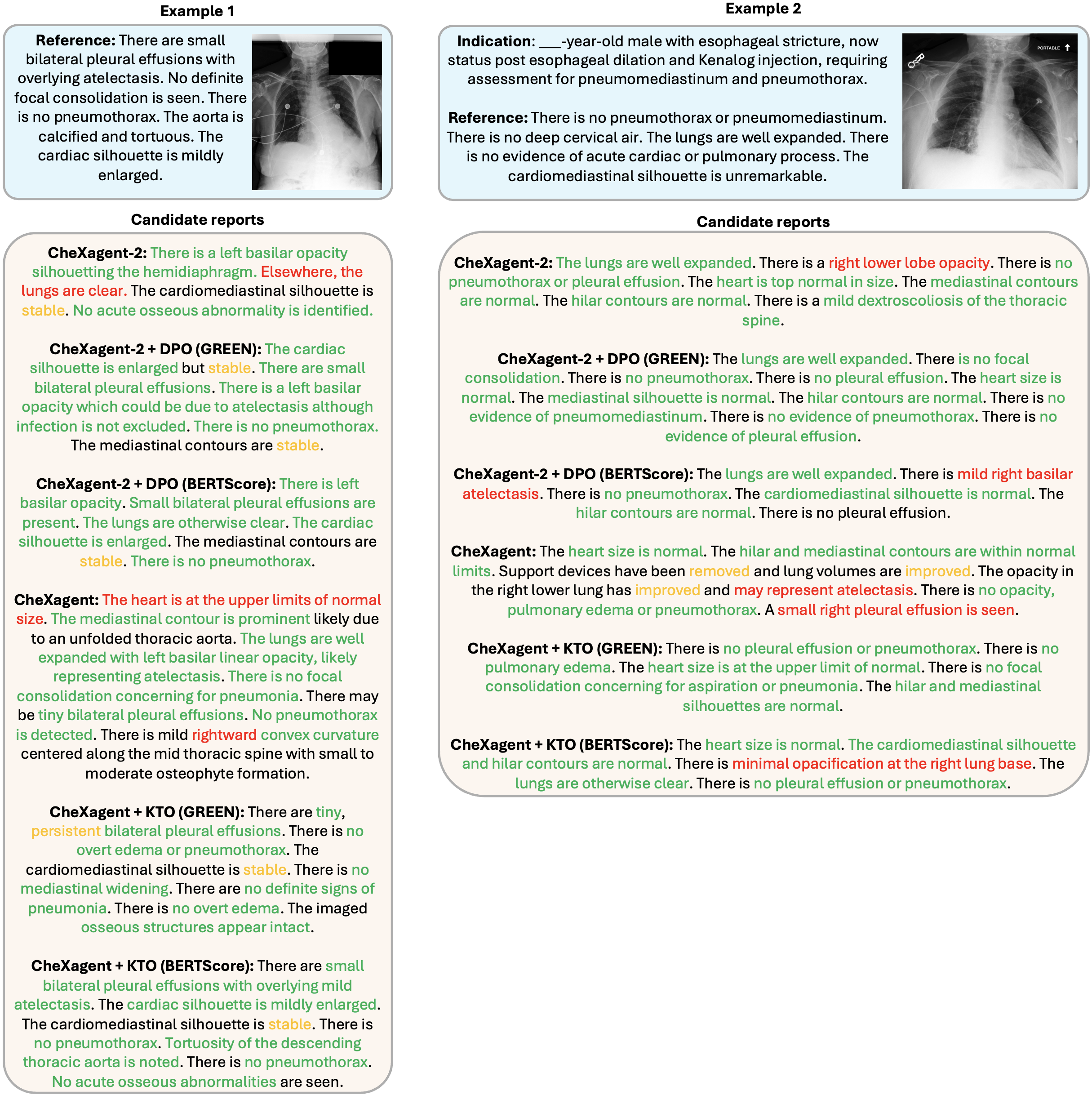}
    \caption{Color-coded candidate reports: Examples 1 and 2. Green and red represent correct and incorrect. Orange refers to prior imaging study.}
    \label{fig:qualitative-appenidx1}
\end{figure}

Example 1 and 2 are available in Fig.~\ref{fig:qualitative-appenidx1}. For example 1, CheXagent-2, despite being a very strong model, contains two errors. First, it incorrectly states that lungs are elsewhere clear---marked in red. Secondly, comparing with the reference report, the detection of small bilateral pleural effusions is missing. None of these errors are made by the aligned versions, using both GREEN and BERTScore as Judge. Moreover, for CheXagent with both GREEN and BERTScore as Judge, there are repetitions. For GREEN ``no overt edema'' is mentioned twice. For BERTScore as Judge, ``no pneumothorax'' is repeated. These types of exact repetitions may be due to verbosity bias. 

Moving on to example 2, we can see that there are exact repetition for CheXagent-2 aligned with DPO. ``There is no pneumothorax'' and ``There is no pleural effusion'' are bot repeated as ``There is no evidence of ...''. Interestingly, such repetition are not present for BERTScore as Judge, nor for the results from CheXagent. Notably, however, both for CheXagent-2 and CheXagent there are errors in the candidate report using BERTScore as Judge for this case. ``Support devices have been removed'' from CheXagent cannot cannot be supported nor refuted definitively. 

\begin{figure}[!htbp]
    \centering
    \includegraphics[width=0.7\linewidth]{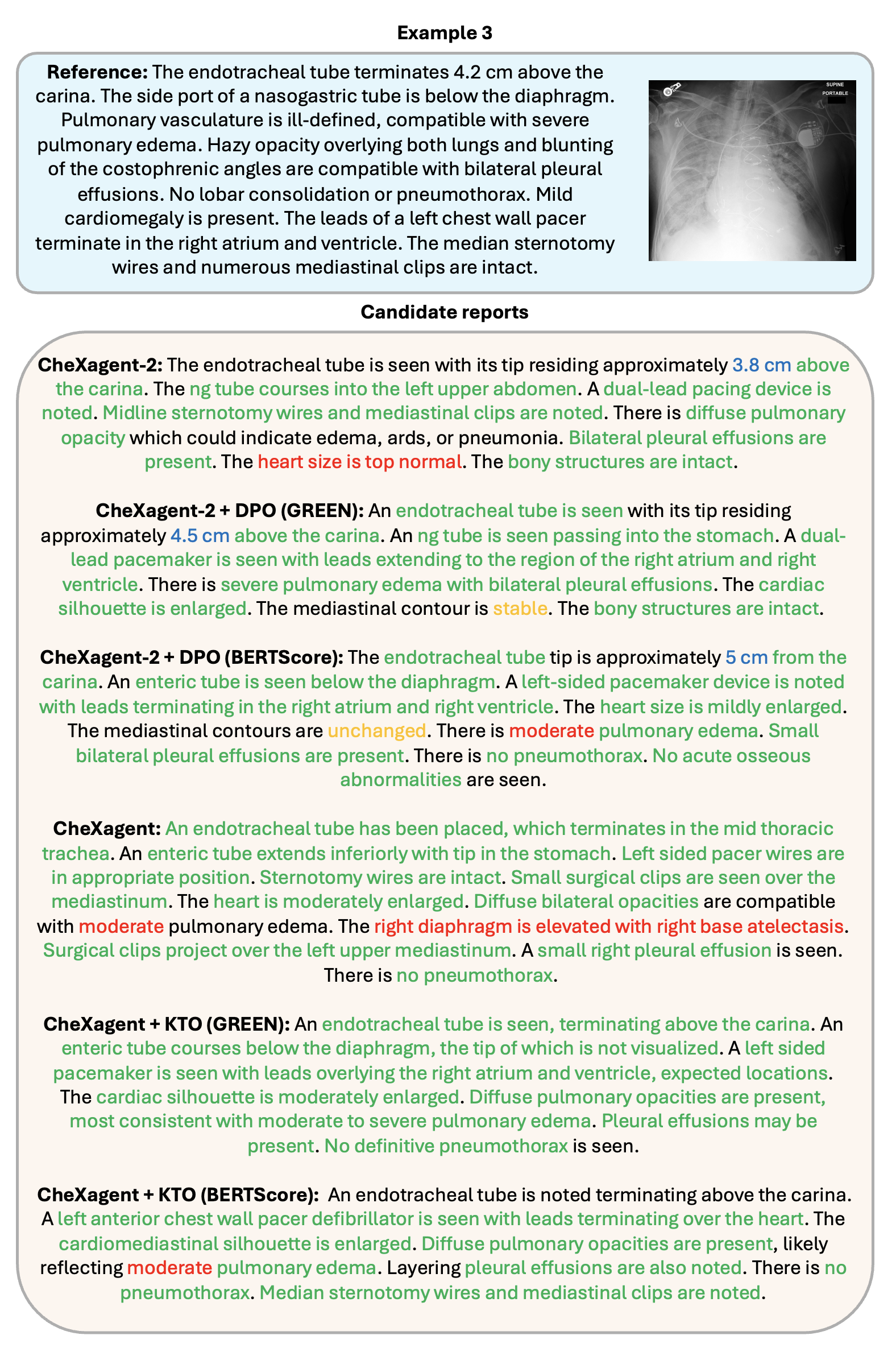}
    \caption{Color-coded candidate reports: Example 3. Green and red represent correct and incorrect. Orange refers to prior imaging study. Blue indicates measurements that have not been verified.}
    \label{fig:qualitative-appenidx2}
\end{figure}

Finally, consider example 3 in Fig.~\ref{fig:qualitative-appenidx2}. This case includes a lot of details, with a reference report significantly longer than the previous two cases. One thing that is very interesting about this case is that the reference report includes a quantitative measurement: the distance between where the endotracheal tube terminates and the carina, expressed in centimeters. Quantitative results are given in some of the candidate reports. However, due to the set up of this reader study (i.e. that we used CXRs in png format) it is not feasible for the radiologist to make measurements and determine distance definitively. To capture this additional feature, we introduce the additional color blue. Takeaways are similar as to the previous two cases. Errors are made by CheXagent-2 and CheXagent, as well as the aligned versions using BERTScore as Judge. For GREEN as Judge, the performance remains strong, with no direct errors. Moreover, cheXagent makes a partial mistake not capture by our color coding scheme in this case as it states that there is ``A small right pleural effusion,'' where it is actually bilateral. 

\end{document}